\documentclass[]{jingdong}

\usepackage[T1]{fontenc}
\usepackage[utf8]{inputenc}
\usepackage{microtype}
\usepackage{graphicx}
\usepackage{url}
\usepackage{booktabs}
\usepackage{multirow}
\usepackage{amsmath}
\usepackage{amssymb}
\usepackage{amsfonts}
\usepackage{amsthm}
\usepackage{mathtools}
\usepackage{nicefrac}
\usepackage{enumitem}
\usepackage{listings}
\usepackage{xcolor}
\usepackage{xspace}
\usepackage{balance}
\usepackage{tikz}
\usepackage{adjustbox}
\usetikzlibrary{arrows.meta,positioning,calc}

\definecolor{EchoAccent}{HTML}{D95C35}

\newcommand{\model}{\textsc{EchoWM}\xspace}

\theoremstyle{plain}

\theoremstyle{definition}

\theoremstyle{remark}

\title{EchoWM: Open and Enterable Omnimodal World Models}
\author[1,3*]{Songchun Zhang}
\author[2,3*]{Yaowei Li}
\author[3,*]{Junhao Zhuang}
\author[4,3*]{Weiyang Jin}
\author[3,5]{Haoyu Wang}
\author[3,6]{Xin Lu}
\author[3]{Yilang Sun}
\author[5]{Shiyi Zhang}
\author[3]{Haoran Li}
\author[3,6]{Xiaoxiao Ma}
\author[3,4]{Yuming Li}
\author[3,5]{Yijun Liu}
\author[3,7]{Yaofeng Su}
\author[3,8]{Yanwen Ma}
\author[3,4]{Haoyu Wu}
\author[3,5]{Zihan Su}
\author[1]{Yue Ma}
\author[9]{Lvmin~Zhang}
\author[3]{Haoyang Huang}
\author[3,\ddagger]{Zeyue Xue}
\author[1,\dagger]{Anyi Rao}
\author[3,\dagger]{Nan Duan}
\affiliation[1]{HKUST}
\affiliation[2]{PKU}
\affiliation[3]{Joy Future Academy, JD}
\affiliation[4]{HKU}
\affiliation[5]{THU}
\affiliation[6]{USTC}
\affiliation[7]{FDU}
\affiliation[8]{Beihang University}
\affiliation[9]{Stanford University}
\contribution[*]{These authors contributed equally.}
\contribution[\dagger]{Corresponding authors.}
\contribution[\ddagger]{Project lead.}

\renewcommand{\titleteaser}{%
  \vspace{-5mm}%
}

\abstract{
We present \model, an omnimodal world model for enterable generative media that responds to continuous navigation while jointly generating 720p video, environmental sound, music and speech.
We organize interaction around camera intent: in first-person scenes, it specifies observer motion, while in third-person scenes, camera--character dynamics are learned from data without view-specific controllers.
Discrete commands and continuous poses are mapped to a shared metric-scale relative 6-DoF trajectory, with dataset-level calibration preserving motion magnitude across heterogeneous data.
To jointly learn audio-visual generation and trajectory control, we construct a complementary data engine and adopt progressive training followed by autoregressive post-training for long-horizon generation.
Extensive evaluations show that \model achieves strong trajectory following and high visual quality on public world-model benchmarks, supporting both first- and third-person interaction across varied subjects, and maintaining synchronized environmental sound and speech over long-horizon generation.

}

\checkdata[Project Page]{\url{https://echo-team-joy-future-academy-jd.github.io/Echo-1.5-Page/wm}}
\checkdata[Code]{\url{https://github.com/jd-opensource/JoyAI-Echo}}

\begin{document}
\maketitle

\begin{center}
    \vspace{-2mm}
    \includegraphics[width=0.85\textwidth]{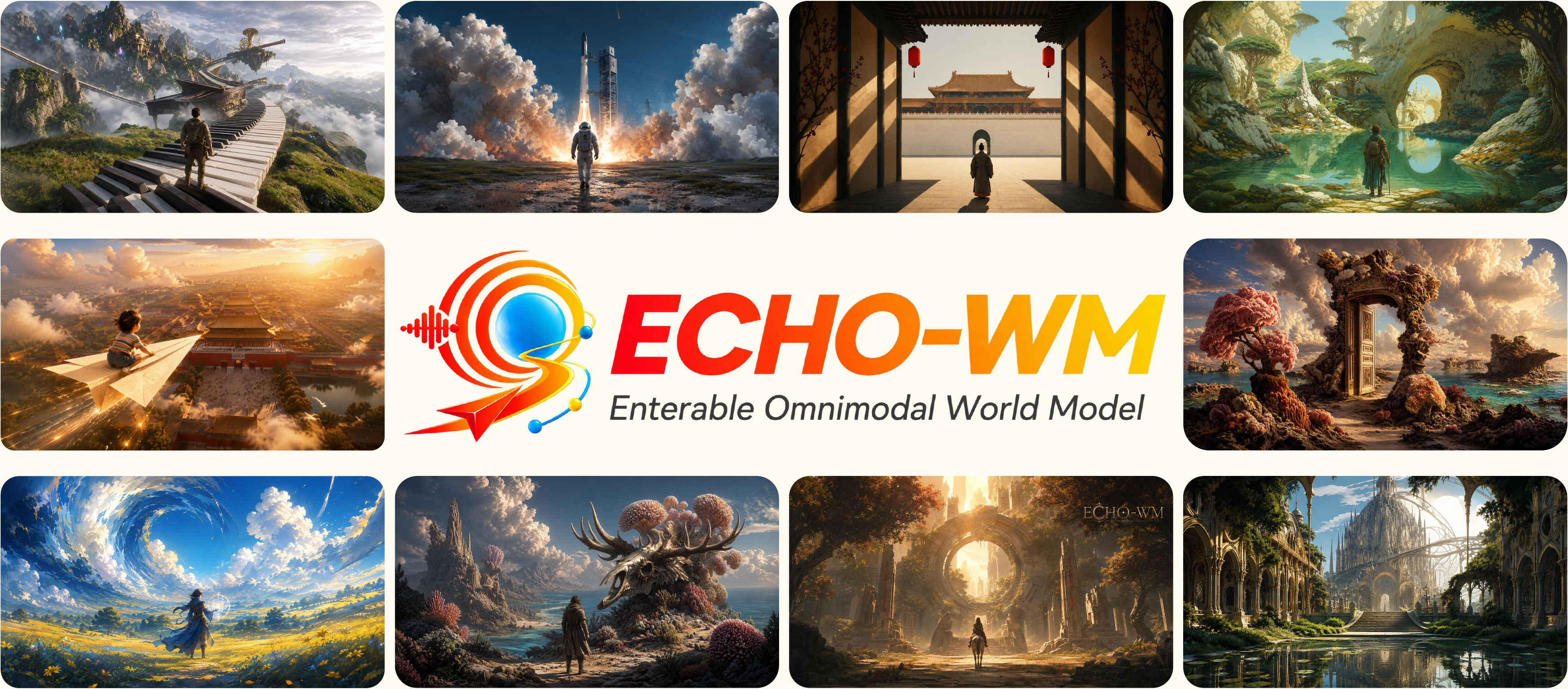}
    \vspace{-2mm}
\end{center}

\begin{center}
    \includegraphics[width=1.0\textwidth]{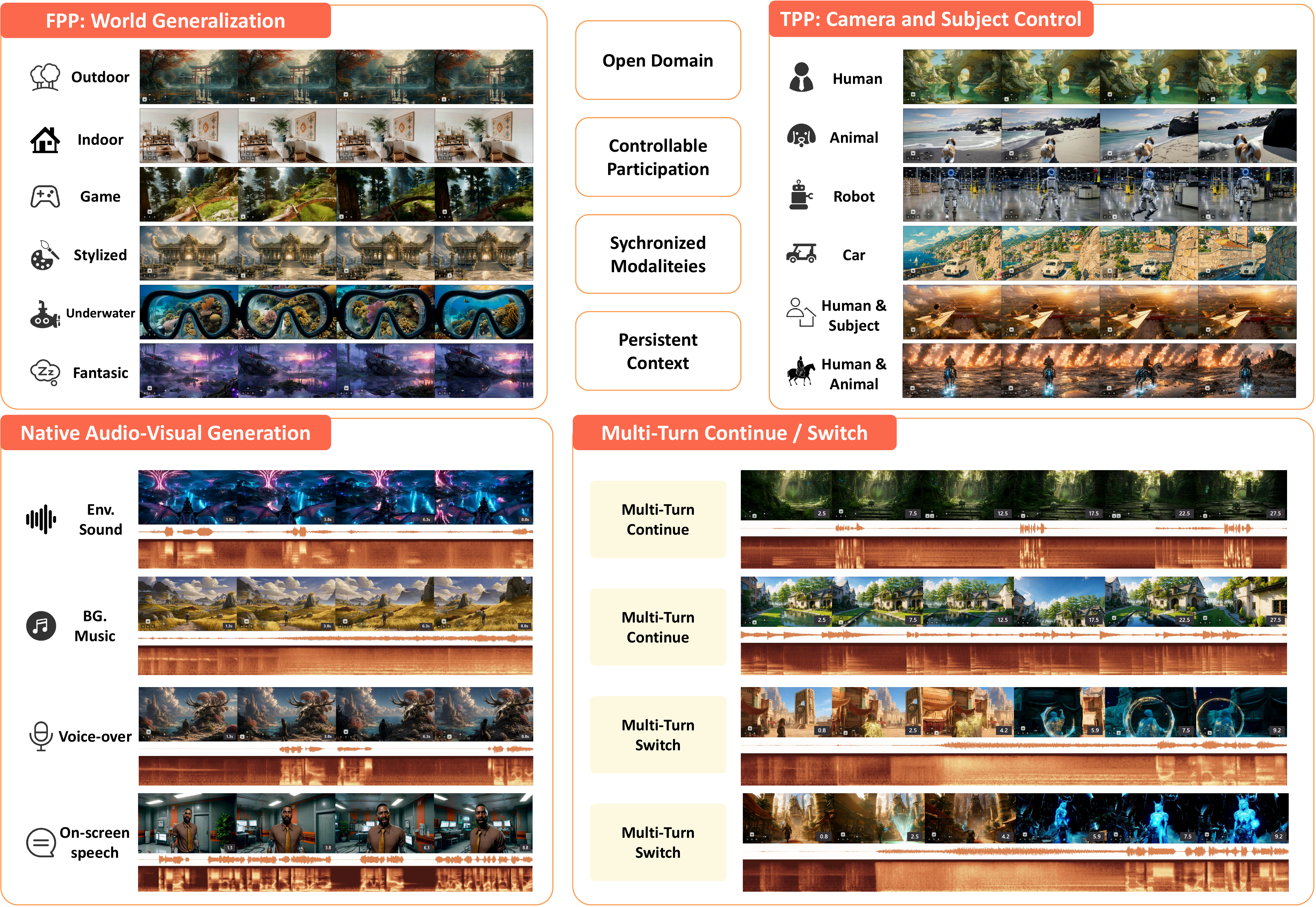}
    \captionof{figure}{\textbf{EchoWM turns audio-visual generation into an enterable world.}
    Given a reference observation, a structured world description, and a user control sequence represented as a relative 6-DoF trajectory, one model supports first- and third-person interaction, synchronized environmental sound and speech, and multi-turn continuation.}
    \label{fig:teaser}
\end{center}

\section{Introduction}
\label{sec:introduction}

Generative video models are rapidly evolving into expressive audio-visual media engines~\citep{polyak2024movie,veo3,hacohen2026ltx2}, yet user interaction remains predominantly prompt-based and passive.
World models invert this relationship by generating future states that respond to continuous user or agent inputs~\citep{unisim,bruce2024genie,gaia1,wang2023drivedreamer,he2024cameractrl,ren2025gen3c,zhu2026sanawm}.
However, most existing systems produce silent visual rollouts or rely on rigid, subject-specific action spaces.
In an interactive world, audio is essential: footsteps and collisions reveal physical feedback, ambient sounds convey spatial dynamics, and speech connects visible subjects to social and narrative context.
While recent work explores joint audio-visual generation in robotics~\citep{nvidia2026cosmos3}, enabling continuous, multimodal participation across general and game scenes remains an open challenge.

To address this gap, we introduce the concept of an enterable omnimodal world: a generated environment that users can continuously traverse with synchronized visual, environmental sound, and speech output.
Here, interaction focuses on navigation and viewpoint evolution representable by continuous trajectories, rather than unrestricted control over arbitrary subject behaviors.
Our key insight is to organize this interaction around camera intent: a subject-agnostic interaction interface grounded in relative 6-DoF trajectories that captures how an observer intends to navigate and view the generated scene.
Operating independently of specific actor dynamics, vehicle controls, or camera-rig constraints, camera intent generalizes seamlessly across diverse subjects and perspectives: in first-person scenes, it directly guides observer motion; in third-person scenes, the same trajectory naturally orchestrates coordinated character locomotion, vehicle displacement, camera tracking, and scene framing.
By learning this observer--subject relationship directly from data, a single interface spans diverse perspectives and domains without requiring viewpoint-specific controllers.

\begin{table}[tbp]
    \centering
    \small
    \newcommand{\capcheck}{\textcolor{EchoAccent}{$\checkmark$}}
    \newcommand{\capnone}{\textcolor{black!30}{--}}
    \setlength{\tabcolsep}{6pt}
    \renewcommand{\arraystretch}{1.12}
    \caption{Reported native capabilities of representative generative video and world models.
    A checkmark denotes explicit native support, whereas ``--'' means that the capability is not reported.
    Pose trajectory and discrete navigation denote continuous and discrete viewpoint-related interaction, respectively.}
    \label{tab:intro-capability-comparison}
    \resizebox{\textwidth}{!}{%
    \begin{tabular}{@{}lccccccc@{}}
        \toprule
        & \multicolumn{1}{c}{\textbf{Output}} &
        \multicolumn{3}{c}{\textbf{Participation}} &
        \multicolumn{3}{c}{\textbf{Audio feedback}} \\
        \cmidrule(lr){2-2}\cmidrule(lr){3-5}\cmidrule(lr){6-8}
        \textbf{Model} & \shortstack{High-res. video} & \shortstack{FPP + TPP} & \shortstack{Continue traj.} & \shortstack{Discrete keyboard} & \shortstack{Env. sound} & \shortstack{Background music} & \shortstack{Spoken speech} \\
        \midrule
        \multicolumn{8}{@{}l}{\textit{Interactive world models}} \\
        LingBot-World~\citep{gao2026lingbot} & \capcheck & \capcheck & \capcheck & \capnone & \capnone & \capnone & \capnone \\
        DreamX-World 1.0~\citep{team2026dreamx} & \capcheck & \capcheck & \capcheck & \capnone & \capnone & \capnone & \capnone \\
        SANA-WM~\citep{zhu2026sanawm} & \capcheck & \capcheck & \capcheck & \capnone & \capnone & \capnone & \capnone \\
        Matrix-Game 3.0~\citep{wang2026matrixgame3} & \capcheck & \capcheck & \capnone & \capcheck & \capnone & \capnone & \capnone \\
        Matrix-Game 3.5~\citep{riemann2026matrixgame35} & \capcheck & \capcheck & \capcheck & \capnone & \capnone & \capnone & \capnone \\
        Genie 3~\citep{deepmind2025genie3} & \capcheck & \capcheck & \capnone & \capcheck & \capnone & \capnone & \capnone \\
        Happy Oyster~\citep{ying2026wbench} & \capcheck & \capcheck & \capnone & \capcheck & \capcheck & \capcheck & \capnone \\
        \midrule
        \multicolumn{8}{@{}l}{\textit{Generative and omnimodal models}} \\
        HunyuanVideo 1.5~\citep{wu2025hunyuanvideo15} & \capcheck & \capnone & \capnone & \capnone & \capnone & \capnone & \capnone \\
        LTX-2.3~\citep{hacohen2026ltx2} & \capcheck & \capnone & \capnone & \capnone & \capcheck & \capcheck & \capcheck \\
        Cosmos 3~\citep{nvidia2026cosmos3} & \capcheck & \capnone & \capnone & \capnone & \capcheck & \capnone & \capnone \\
        \midrule
        \textbf{\model} & \capcheck & \capcheck & \capcheck & \capcheck & \capcheck & \capcheck & \capcheck \\
        \bottomrule
    \end{tabular}%
    }
\end{table}

Realizing an enterable omnimodal world under this unified interface requires resolving three coupled geometric and distributional mismatches, which we address through three dedicated technical designs:
\begin{itemize}[leftmargin=*, topsep=2pt, itemsep=2pt]
    \item \textbf{Architecture (Unifying Viewpoint Semantics \& Scale):} Heterogeneous video sources differ fundamentally in metric scale and control semantics; naive normalization can erase displacement magnitude or degrade control consistency. We design a unified architecture that maps both discrete commands and continuous poses to a shared relative 6-DoF trajectory with a fixed dataset-level metric scale, while encoding pairwise camera geometry via relative UCPE. Combined with viewpoint context, the learned world prior interprets this intent as either first-person observer motion or third-person camera--subject evolution.
    \item \textbf{Data (Bridging AV Richness \& Control Precision):} Real-world videos offer rich acoustics but messy motion, whereas simulations provide precise geometry with simple semantics. We build a world-data engine from four complementary sources (internal gameplay, web gameplay, UE simulation, and general web video) and structure them into AV-rich, control-clean, and balanced high-quality mixtures, allowing the model to jointly learn multimodal quality and control responsiveness without relying on a single ideal data source.
    \item \textbf{Training (Decoupling Multimodal Quality \& Control Responsiveness):} Unifying audio-visual synthesis with precise control induces severe training distribution conflicts. We therefore use a progressive four-stage curriculum in which Audio-Visual Continued Pretraining (AV-CPT) establishes visual dynamics and acoustic priors, Action Fine-Tuning (Action-SFT) isolates control sensitivity by freezing the backbone and training only a lightweight trajectory pathway, Joint Fine-Tuning (Joint-FT) consolidates both capabilities with a low learning rate on balanced data, and Autoregressive Post-Training adapts the model to self-generated context for long-horizon inference.
\end{itemize}

We instantiate these principles in \model, an omnimodal world model for enterable generative media.
Given a reference observation and a textual world description, \model responds to continuous navigation by jointly generating 720p video, environmental audio, and speech with multi-turn continuation.
As shown in Table~\ref{tab:intro-capability-comparison} and Figure~\ref{fig:teaser}, \model covers real indoor/outdoor scenes, stylized game worlds, artistic content, and diverse subjects.
Quantitative evaluations demonstrate state-of-the-art performance across public benchmarks: \model ranks first on WBench Navigation and achieves top-tier visual quality on SANA-WM-Bench across both simple and challenging trajectory splits, while maintaining robust visual and control consistency over long-horizon rollouts.

Our main contributions are summarized as follows:
\begin{itemize}[leftmargin=*, topsep=2pt, itemsep=3pt]

    \item \textbf{Enterable Omnimodal World Modeling.}
    We formulate an enterable generative-media setting that combines continuous
    user navigation with native joint generation of video, environmental sound,
    and speech.
    The formulation supports both first- and third-person interaction and
    extends beyond a single rollout through synchronized multi-turn continuation.

    \item \textbf{Unified Camera-Intent Interface.}
    We organize navigation around camera intent and represent heterogeneous
    discrete commands and continuous camera poses with a shared relative 6-DoF
    trajectory.
    Dataset-level metric calibration and relative UCPE provide consistent
    geometric conditioning across data sources, while the learned
    observer--subject relationship allows the same interface to support
    first-person observer motion and third-person camera--character evolution
    without view-specific controllers or camera rigs.

    \item \textbf{Capability-Aligned World Data Engine.}
    We build a scalable data engine from internally collected gameplay,
    human-played Internet gameplay, Unreal Engine simulation, and general
    Internet video, combining complementary audio-visual and geometric signals.
    The processed data are organized into AV-rich, control-clean, and balanced
    high-quality mixtures so that audio-visual generation and trajectory
    following are trained from the subsets in which their supervision is most
    reliable.

    \item \textbf{Progressive Training.}
    We introduce a progressive training strategy that first adapts joint
    audio-visual generation, then learns trajectory conditioning with the
    backbone frozen, and subsequently fine-tunes both components together.
    Autoregressive post-training further exposes the model to self-generated
    audio-visual histories, enabling stable long-horizon and multi-turn
    interactive generation.

\end{itemize}


\section{Related Work}
\label{sec:related-work}

\subsection{Interactive Generative World Models}

Generative world models have increasingly moved from compact latent dynamics
for imagination and control toward domain-specific action-conditioned simulation
and open-ended interactive video generation~\citep{ha2018world,hafner2018planet,hafner2019dream}.
Early systems such as UniSim and Genie learn visual futures conditioned on
agent actions~\citep{unisim,bruce2024genie}, while recent game-oriented models
demonstrate real-time or playable neural simulation~\citep{li2025hunyuangamecraft,mao2025yume,savva2026solaris}.
More recent systems substantially broaden visual fidelity, domain coverage,
and interaction horizon~\citep{cosmos_v1,gaia1,wang2023drivedreamer,robodreamer}.
Genie~3 supports real-time exploration of diverse 720p environments~\citep{deepmind2025genie3};
WorldPlay and Matrix-Game focus on streaming generation and long-term
geometric consistency~\citep{sun2025worldplay,he2025matrixgame2,wang2026matrixgame3};
LingBot-World, SANA-WM, and DreamX-World further
extend controllable world generation toward broader visual domains and
longer trajectories~\citep{gao2026lingbot,zhu2026sanawm,team2026dreamx}.
Most of these models nevertheless remain primarily visual world simulators,
with interaction interfaces often tailored to specific domains, environments, or control schemes. Our work instead focuses on the complementary problem of building a single
camera-centric interaction interface that spans first- and third-person
navigation while natively generating video, environmental sound, and speech
across general and game scenes.

\subsection{Action and Camera Conditioning for Interactive Generation}

Interactive generative models commonly expose control through either discrete
actions or continuous camera geometry.
Action-conditioned world models such as ABot-World and Hunyuan-GameCraft
condition future generation on keyboard or controller inputs, supporting
navigation and character control over time~\citep{jiang2026abotworld,li2025hunyuangamecraft}.
Such interfaces are intuitive, but their action semantics are typically tied
to a particular controller or environment.
Camera-controllable video models instead condition generation on poses or
trajectories.
A common approach represents camera rays with Pl\"ucker coordinates:
CameraCtrl~\citep{he2024cameractrl}, for example, processes dense Pl\"ucker
features with a dedicated camera encoder, while the LingBot-World family
injects Pl\"ucker-based camera conditions through feature modulation such as
AdaLN~\citep{gao2026lingbot}.
MotionCtrl, CamCo, and GEN3C extend camera control to flexible motion and
3D-consistent generation~\citep{wang2024motionctrl,xu2024camco,ren2025gen3c}.
More recent methods such as UCPE~\citep{zhang2025ucpe} incorporate
relative camera geometry directly into attention, reducing reliance on a
standalone camera encoder.
These lines of work differ in both user-facing control spaces and internal geometric representations.
EchoWM maps discrete navigation commands and continuous metric poses to a shared relative 6-DoF trajectory, using the same geometric interface across first- and third-person interaction.

\subsection{Joint Audio-Visual and Omnimodal Generation}

Audio generation for video has traditionally been studied as a
video-to-audio problem, where sound is synthesized after the visual stream~\citep{luo2023difffoley,zhang2024foleycrafter,mmaudio}.
More recent generative models instead produce visual and acoustic content
jointly, including Movie Gen, Veo~3, SyncFlow, and LTX-based audio-visual
diffusion models~\citep{polyak2024movie,veo3,liu2024syncflow,hacohen2026ltx2}.
OmniForcing further converts joint audio-visual diffusion into a streaming
autoregressive generator while preserving cross-modal synchronization~\citep{su2026omniforcing}.
In parallel, omnimodal world models such as Cosmos~3 incorporate video,
audio, language, and action for embodied-agent and Physical-AI settings~\citep{nvidia2026cosmos3}.
These advances establish strong native audio-visual generation, but most do not expose continuous geometric control over the generated world.
EchoWM focuses on combining native environmental sound and speech with a
shared trajectory interface for first- and third-person interaction across
general and game-oriented generative media.

\subsection{Autoregressive and Long-Horizon Video Generation}
Autoregressive video diffusion replaces bidirectional attention with
causal computation.
Diffusion Forcing adapts full-sequence generators to causal
prediction~\citep{chen2024diffusion}, and Self Forcing trains on
self-generated rollouts~\citep{huang2026self}.
Long context is then treated as memory: FramePack bounds the token
budget, and Longlive keeps an attention sink of early
frames~\citep{zhang2025framepack,yang2025longlive}.
Interactive worlds go further with explicit retrieval.
WorldMem and Context-as-Memory index past views by pose or field of
view~\citep{xiao2025worldmem,yu2025contextasmemory};
Infinite-World, WorldPlay, and Matrix-Game~3.0 carry hierarchical or
camera-aligned memories across long
rollouts~\citep{wu2026infiniteworld,sun2025worldplay,wang2026matrixgame3}.
OmniForcing and Self Gradient Forcing extend this lineage to
synchronized audio-video streams and native long-video
extrapolation~\citep{su2026omniforcing,zhuang2026sgf,savva2026solaris}.
We apply a similar causal post-training strategy to trajectory-conditioned joint audio-visual world generation. Unlike explicit retrieval, which incurs additional computation for searching and matching against long-term history, the generator maintains a bounded sink-plus-FIFO history and writes a compact state from past chunks only, allowing long-horizon interaction to reuse native memory even after recent tokens leave the active window.
\section{World Data Engine}
\label{sec:data}

\begin{figure}[ht]
    \centering
    \includegraphics[width=1.0\linewidth]{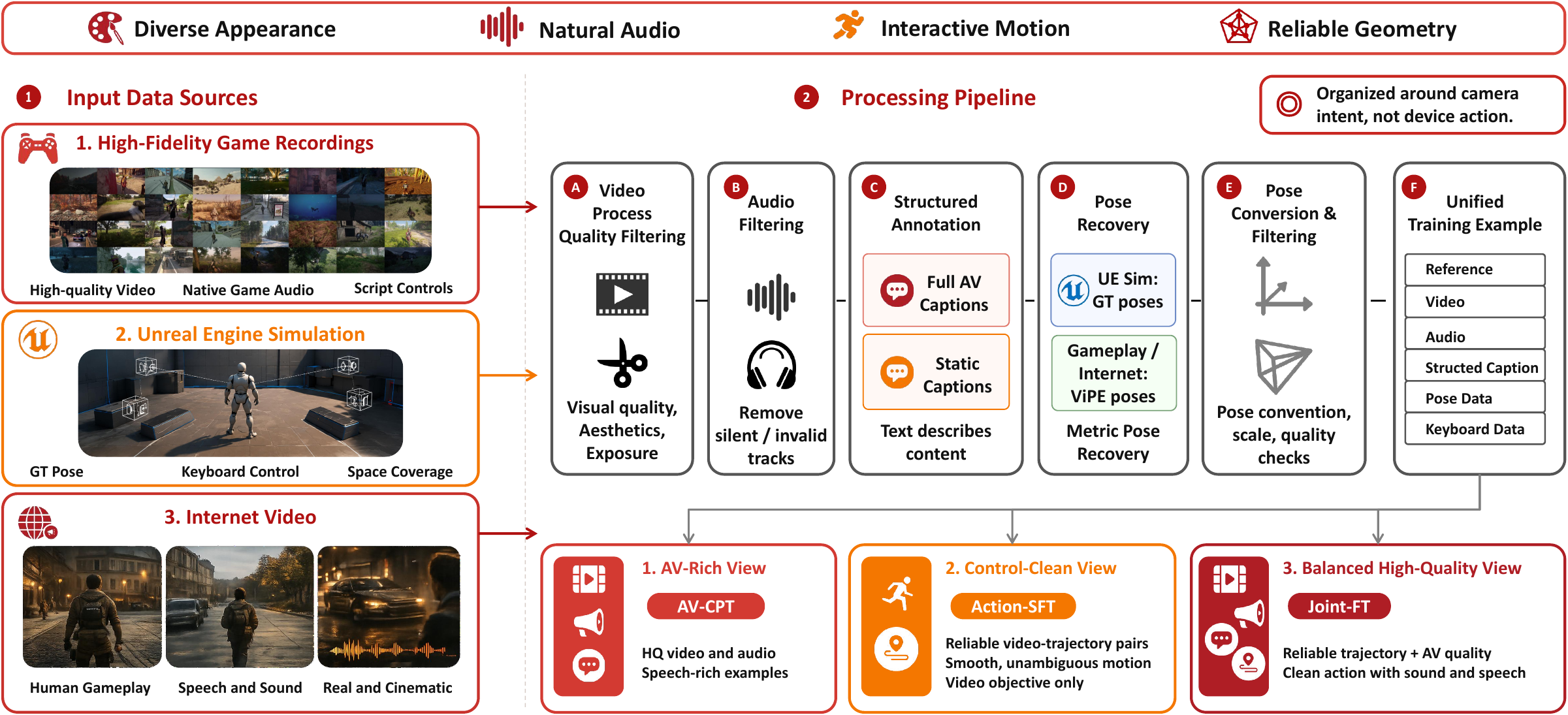}
    \caption{\textbf{World data engine.}
    We combine internally collected game recordings, human-played Internet game recordings, Unreal Engine renders, and general Internet video to complement action logs, natural audio, interactive motion, reliable poses, and cross-domain appearance.
    A shared pipeline constructs synchronized clips, filters and annotates video and audio, recovers reliable metric poses, and forms AV-rich, control-clean, and balanced high-quality mixtures for AV-CPT, Action-SFT, and Joint-FT, respectively.}
    \label{fig:data-pipeline}
\end{figure}

Training an enterable omnimodal world requires four properties that rarely coexist in a single data source: diverse appearance, natural audio, interactive motion, and reliable camera geometry.
We therefore combine four complementary sources: internally collected game recordings, human-played Internet game recordings, Unreal Engine (UE) simulation, and general Internet video.
Together, these sources provide complementary supervision for audio-visual world modeling and trajectory-conditioned interaction.
Internally collected gameplay provides controlled interaction, native audio, and
action logs; human-played gameplay contributes natural control timing and richer
speech; UE provides ground-truth metric geometry and controlled motion coverage;
and general Internet video broadens appearance and acoustic diversity beyond
game-rendered worlds.
We process them through two complementary paths: an audio-visual path that
prioritizes appearance, speech, and environmental sound, and a geometry path
that preserves long-range temporal continuity for metric camera recovery.
The two paths are joined after processing into temporally aligned examples with
structured audio-visual annotations and reliable metric camera trajectories
wherever geometry can be recovered.
The resulting examples are organized into three stage-aligned mixtures for AV-CPT, Action-SFT, and Joint-FT.
\Cref{tab:data-source-comparison} summarizes the complementary signals and limitations of the four sources, while \cref{fig:data-pipeline} illustrates the overall source-to-mixture pipeline.

\begin{table}[tbp]
    \centering
    \footnotesize
    \newcommand{\srcdoton}{\textcolor{EchoAccent}{\raisebox{0.12ex}{\scalebox{0.68}{$\bullet$}}}}
    \newcommand{\srcdotoff}{\textcolor{black!18}{\raisebox{0.12ex}{\scalebox{0.68}{$\circ$}}}}
    \newcommand{\srccheck}{\makebox[2.4em][c]{\textcolor{EchoAccent}{\raisebox{0.05ex}{$\checkmark$}}}}
    \newcommand{\srcHigh}{\makebox[2.6em][c]{\srcdoton\hspace{0.08em}\srcdoton\hspace{0.08em}\srcdoton}}
    \newcommand{\srcMid}{\makebox[2.6em][c]{\srcdoton\hspace{0.08em}\srcdoton\hspace{0.08em}\srcdotoff}}
    \newcommand{\srcLow}{\makebox[2.6em][c]{\srcdoton\hspace{0.08em}\srcdotoff\hspace{0.08em}\srcdotoff}}
    \newcommand{\srcnone}{\makebox[2.4em][c]{\textcolor{black!30}{--}}}
    \setlength{\tabcolsep}{2.3pt}
    \renewcommand{\arraystretch}{1.20}
    \caption{Complementary strengths of the four data sources.
    Filled versus hollow dots encode relative source strength; checkmarks indicate availability; dashes indicate absence.
    Scene dynamics refers to subject and environment motion beyond the commanded camera or character trajectory.}
    \label{tab:data-source-comparison}
    \resizebox{\textwidth}{!}{%
    \begin{tabular}{@{}lccccccccc@{}}
        \toprule
        & \multicolumn{4}{c}{\textbf{Audio-visual content}} &
        \multicolumn{3}{c}{\textbf{Action signal}} &
        \multicolumn{2}{c}{\textbf{Coverage}} \\
        \cmidrule(lr){2-5}\cmidrule(lr){6-8}\cmidrule(lr){9-10}
        \multirow{2}{*}{\textbf{Source}} & Visual & Audio & \multirow{2}{*}{Speech} & Scene & Action & \multirow{2}{*}{Logs} & Pose & Human & Non-game \\
        & quality & richness & & dynamics & clarity & & reliability & operation & domain \\
        \midrule
        Collected gameplay & \srcHigh & \srcHigh & \srccheck & \srcHigh & \srcHigh & \srccheck & \srcMid & \srcnone & \srcnone \\
        \addlinespace[1.5pt]
        Internet gameplay & \srcMid & \srcHigh & \srccheck & \srcHigh & \srcMid & \srcnone & \srcMid & \srccheck & \srcnone \\
        \addlinespace[1.5pt]
        UE renders & \srcMid & \srcnone & \srcnone & \srcnone & \srcHigh & \srccheck & \srcHigh & \srcnone & \srcnone \\
        \addlinespace[1.5pt]
        General Internet video & \srcHigh & \srcHigh & \srccheck & \srcHigh & \srcLow & \srcnone & \srcLow & \srcnone & \srccheck \\
        \bottomrule
    \end{tabular}%
    }
\end{table}

\begin{figure}[t!]
    \centering
    \includegraphics[width=\textwidth]{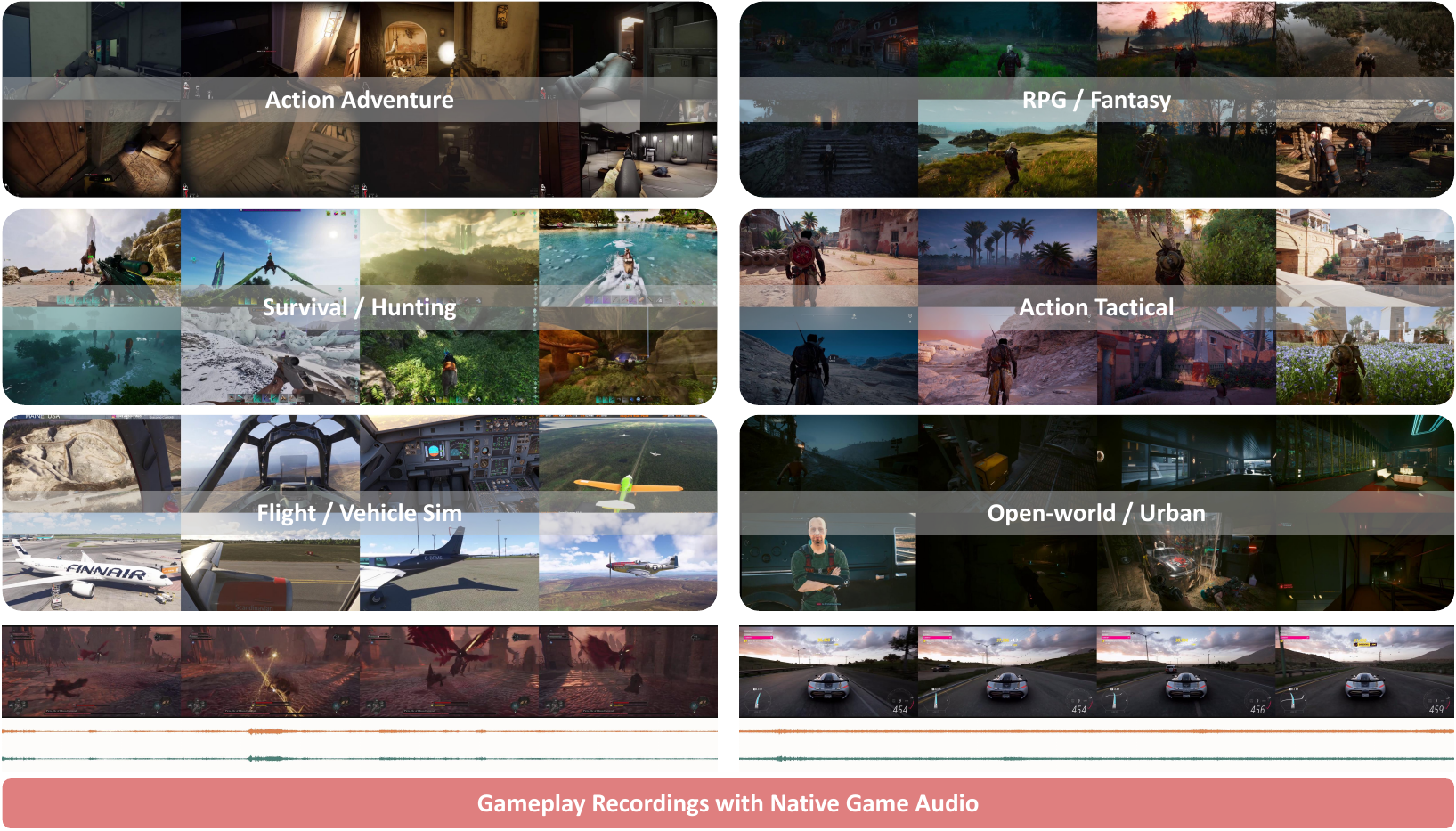}
    \caption{\textbf{Gameplay recordings with native game audio.}
    Representative frames from the internally collected gameplay corpus span
    six activity categories, with synchronized native-game audio waveforms
    shown below the visual sequences.}
    \label{fig:data-vis}
\end{figure}

\begin{figure}[t!]
    \centering
    \includegraphics[width=\textwidth]{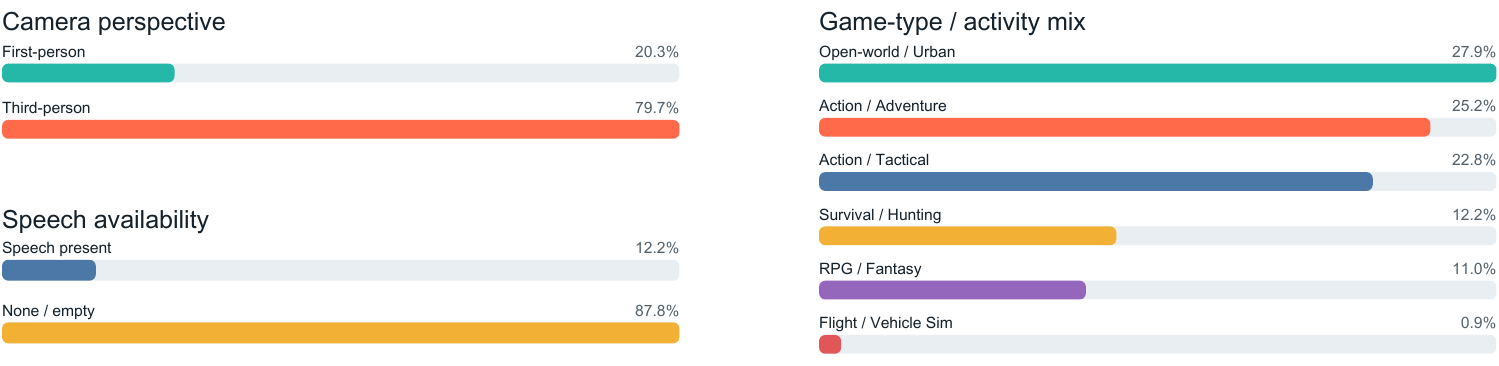}
    \caption{\textbf{Game-video dataset statistics.}
    The combined training corpora contain predominantly third-person clips,
    a broad mix of game activities, and a smaller speech-present subset.
    Percentages are normalized within the reported collection.}
    \label{fig:ltx-dataset-statistics}
\end{figure}

\subsection{Complementary Data Sources}

\noindent\textbf{Internally collected gameplay.}
Our internally collected game recordings form a major source of trajectory-conditioned data and span first- and third-person play across open-world exploration, urban driving, and equestrian traversal.
Programmatic scripts control the characters, providing explicit control inputs together with cleaner action--observation correspondence than uncontrolled Internet video.
A ReShade-based plug-in suppresses heads-up displays and interface overlays before capture, preserving the original field of view without post-hoc cropping.
We synchronously record high-resolution RGB video, the native stereo game mix, script-issued action logs, and environment metadata.
The native mix retains dialogue, music, ambience, and interaction sounds such as footsteps, vehicles, and collisions.
Representative visual sequences and synchronized native-game audio waveforms are shown in \cref{fig:data-vis}.

Importantly, native action logs and realized camera trajectories capture complementary aspects of interaction.
An issued control does not necessarily induce the corresponding camera displacement.
For example, a forward command may produce little or no translation when the character is blocked by geometry, constrained by terrain, or affected by game physics.
The action stream therefore records control intent, whereas the camera trajectory records the motion actually realized by the environment.
We preserve both signals in the dataset, while using the recovered camera trajectory as the model's unified motion condition.
Because camera poses are not exported from these games, we recover metric trajectories from video as described in \cref{sec:metric-pose-recovery}.

\noindent\textbf{Human-played Internet gameplay.}
Human-played Internet game recordings complement internally collected gameplay with less scripted and more context-dependent interaction.
Although source-native action logs are unavailable, these recordings exhibit natural control timing and contain reactions to scene events, in-game dialogue, player commentary, and more diverse human speech.
Continuous clips with reliable recovered camera trajectories contribute to trajectory-conditioned training, while the broader pool strengthens audio-visual and speech modeling.

\noindent\textbf{Unreal Engine simulation.}
UE complements game recordings with ground-truth metric poses, synchronized action logs, and controlled translation and rotation coverage, but does not provide natural audio.
We drive characters in the Play-In-Editor world using local-frame navigation commands and engine physics rather than navigation-mesh paths, allowing realized motion to reflect collisions and environmental constraints.
When a character encounters an obstacle, the runtime controller adjusts its heading and continues the rollout.

Each character carries one head-mounted first-person camera and four character-relative third-person cameras that remain oriented toward and move with the character.
At every step, we export synchronized frames, keyboard states, ground-truth
camera poses and intrinsics, and scene metadata for all five views.
Training randomly selects one view together with its corresponding pose and viewpoint metadata from each rollout.
Both signals are retained, but the current model is conditioned only on the
realized camera trajectory.
The capture rig itself is never exposed to the model.

\noindent\textbf{General Internet video.}
General Internet video broadens the training distribution beyond game-rendered worlds with diverse real-world and cinematic visual and acoustic content.
It contributes realistic appearance, environmental sound, speech, richer composition, and camera--subject relationships such as tracking, orbiting, and reframing.
Because these clips provide neither source-native action logs nor standardized navigation, they primarily expand the audio-visual prior and improve cross-domain generalization.
A smaller subset of temporally continuous clips that passes metric pose recovery and motion-quality filtering is additionally included in trajectory-conditioned training.
The resulting game-video distribution is summarized in \cref{fig:ltx-dataset-statistics}.

\subsection{Temporal Segmentation and Canonicalization}

We process the data through separate audio-visual and geometry paths because
the two objectives impose different requirements on temporal continuity.

\noindent\textbf{Audio-visual path.}
For audio-visual training, shot-boundary detection first isolates continuous spans in Internet and cinematic footage.
AV-eligible sources are then sliced directly into short model-ready clips, since learning visual appearance, speech, and environmental sound does not require long-range geometric reconstruction.

\noindent\textbf{Geometry path.}
For non-UE data used in trajectory-conditioned training, we instead preserve
continuous windows of approximately one minute before producing the final short
clips. Reconstructing the long window first ensures that neighboring training
clips inherit a shared geometric reconstruction rather than being estimated in
separate coordinate systems. The longer window also provides broader temporal
support and more persistent correspondences for geometric optimization.
Metric pose recovery and trajectory-level quality control are therefore performed on these long windows before they are sliced into model-ready clips.

\noindent\textbf{Canonicalization.}
All retained media are re-encoded into canonical video and audio formats for stable decoding and batched training, while preserving stereo channels when present.

\subsection{Metric Pose Recovery and Quality Filtering}
\label{sec:metric-pose-recovery}

Trajectory-conditioned training requires camera trajectories that are both temporally consistent and metrically scaled across heterogeneous data sources.
UE provides camera-to-world poses directly in metric units, whereas the remaining trajectory-eligible sources require pose recovery from video.

\subsubsection{Long-Sequence Metric Pose Recovery}

Following the robust annotation principle of SANA-WM~\citep{zhu2026sanawm}, we separate pose estimation from subsequent trajectory quality control rather than directly exposing raw estimator outputs to training.

We use ViPE~\citep{huang2025vipe} as the optimization framework and combine
VGGT-Omega~\cite{wang2026vggt} with MoGe-2 as complementary geometric
backends. VGGT-Omega provides temporally coherent long-sequence geometry,
while MoGe-2 supplies per-frame metric-depth constraints for metric-scale
geometric optimization. ViPE jointly integrates these cues to optimize
temporally consistent metric-scale camera poses and camera intrinsics.
We additionally adapt ViPE to support per-frame intrinsic optimization to
accommodate zoom, stabilization, cropping, and imperfectly known intrinsics in
Internet videos. Together, these components produce metric camera trajectories
and aligned intrinsics for long videos in the wild.

We perform pose recovery on approximately one-minute continuous windows rather
than independently on the final short training clips. This shared reconstruction
keeps neighboring clips in a common geometric coordinate system while providing
broader temporal support for geometric optimization.

\subsubsection{Temporal Densification and Alignment}

Because processing every frame of a long sequence is computationally expensive, we temporally subsample each window before pose estimation.
The recovered sparse trajectory is then aligned back to the original video timestamps using spherical linear interpolation (SLERP) for camera rotations and linear interpolation for translations.
This interpolation is used only to restore frame-aligned trajectory annotations at the original timestamps; the underlying video observations are not synthesized or modified.
The resulting dense trajectories and intrinsics are converted to a common
camera-to-world convention, axis orientation, metric length unit, and temporal
sampling rate. Each long sequence is then sliced into short
trajectory-conditioned examples using identical temporal boundaries for video,
available stereo audio, pose, intrinsics, and native control streams.
All timestamps are re-indexed relative to the beginning of each short clip.
UE rollouts bypass pose reconstruction and interpolation because their camera poses are available directly from the renderer, but follow the same convention conversion, temporal alignment, and final clip-construction pipeline.

\begin{figure}[t!]
    \centering
    \includegraphics[width=0.9\textwidth]{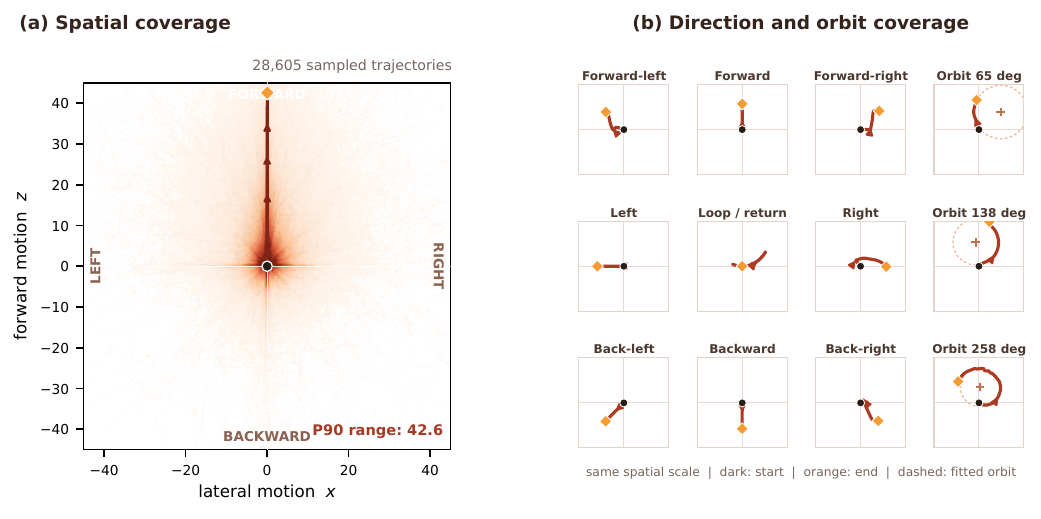}
    \caption{\textbf{Trajectory coverage of the control-clean mixture.}
    (a) Aggregate lateral--forward motion coverage over 28,605 sampled trajectories; the marked P90 range summarizes the robust spatial extent.
    (b) Representative straight, diagonal, return-loop, and orbit trajectories shown at a common spatial scale.
    Black and orange markers denote trajectory start and end points, respectively, and dashed circles show fitted orbits.}
    \label{fig:control-clean-trajectory-coverage}
\end{figure}

\subsubsection{Trajectory Quality Filtering}

We apply trajectory-level quality control before reconstructed sequences are admitted to trajectory-conditioned training.
Rather than relying on a single estimator score, we evaluate three complementary aspects of trajectory quality:
(1) reconstruction reliability and intrinsic-calibration stability;
(2) temporal consistency, including excessive frame-to-frame translation or rotation jitter and abrupt orientation changes; and
(3) motion plausibility, including implausible displacement, degenerate
trajectories, and extreme scale outliers.
Sequences that fail these checks are removed before final short-clip construction.

Ground-truth UE trajectories undergo the same coordinate-convention conversion and motion-range checks, while their original metric values are preserved exactly.
The retained poses are subsequently converted into the relative trajectory representation and dataset-level translation normalization described in \cref{sec:metric-trajectory-interface,sec:pose-normalization}.

\subsection{Audio-Visual Filtering and Structured Annotation}

\noindent\textbf{Audio-visual filtering.}
We remove visually degenerate clips using perceptual quality, aesthetic
quality, exposure, and brightness criteria.
For sources with audio, we additionally reject invalid tracks and clips with
unusable loudness statistics, while separately identifying speech-containing
examples for stage-wise sampling.
Sources without an audio stream, such as UE renders, are treated as
missing-audio examples rather than semantic silence and are not used as
positive supervision for audio content.

\noindent\textbf{Structured audio-visual annotation.}
We use Qwen3-Omni~\citep{xu2025qwen3omni} and Gemini-3-Pro in the annotation
pipeline to process synchronized video and audio and produce structured
descriptions of both persistent scene properties and time-varying audio-visual
events. Their outputs are consolidated into the common structured schema below.
The source records may use field-specific names such as
\texttt{static\_scene\_caption}, \texttt{character},
\texttt{perspective}, \texttt{narrative\_caption}, and \texttt{sounds}; we
normalize these to the canonical fields \textbf{scene}, \textbf{subject},
\textbf{viewpoint}, \textbf{narrative}, and \textbf{sound}, respectively.
Generation metadata and the raw tagged response are retained for provenance
but are not exposed as semantic conditioning fields.

\begin{itemize}[leftmargin=*,itemsep=3pt,topsep=2pt]
    \item \textbf{Persistent context.}
    The \textbf{scene} field describes the environment, spatial layout,
    persistent geometry, and ambient state while excluding the controlled
    subject and individual actions. The \textbf{style} field captures
    rendering medium, realism, lighting, art direction, and color treatment.
    The \textbf{viewpoint} field specifies the initial first- or third-person
    observer--subject relationship without describing subsequent camera
    motion. The \textbf{subject} field describes the primary subject's
    identity, appearance, clothing, and equipment without encoding subsequent
    motion or interaction.

    \item \textbf{Temporal audio-visual events.}
    The \textbf{narrative} field records chronological scene evolution,
    including subject behavior, interactions, and camera motion. The
    \textbf{speech} field records spoken content and, when identifiable, the
    speaker role, distinguishing visible speakers, off-screen sources, and
    player commentary. The \textbf{sound} field describes non-speech acoustics,
    including environmental ambience, music, and interaction sound effects.
\end{itemize}

\noindent\textbf{Preventing textual motion leakage.}
For trajectory-conditioned training, we exclude the narrative field because
it may explicitly describe subject motion and camera evolution.
Otherwise, the model could infer the target motion from text rather than
from the supplied trajectory.
During Action-SFT, we therefore retain the scene, style, viewpoint, and
subject fields while removing the narrative, speech, and sound descriptions;
the static fields and the reference observation define the initial scene,
leaving the trajectory as the only condition that specifies subsequent camera
evolution. For AV-CPT, the full annotation is retained, including narrative,
speech, and sound. For Joint-FT, the static fields are combined with speech and
sound, while narrative remains excluded. This separation prevents explicit
camera-motion leakage while preserving audio conditioning in the stages that
optimize the audio-visual backbone.

\subsection{Stage-Aligned Data Mixtures}

After processing, each example retains the subset of synchronized signals
available from its source, including video, structured text, viewpoint
metadata, camera intrinsics, audio when present, metric camera trajectory when
reliable, and native action logs when available.
Action logs and camera trajectories are intentionally preserved as distinct signals: the former record issued control intent, whereas the latter describe motion actually realized under scene geometry and environment dynamics.
The current model uses camera trajectories as its unified motion condition, while native action streams remain available for alignment, data analysis, and future interaction modeling.

Rather than sampling uniformly from the processed corpus, we construct three mixtures aligned with the three stages of progressive training.
The curriculum progresses from broad audio-visual modeling, to clean trajectory control, and finally to joint consolidation.
The AV path supplies the broad AV-rich pool, whereas the geometry path filters
for reliable trajectories and supplies the control-clean pool; their
high-quality intersection forms the balanced mixture for Joint-FT.

\begin{itemize}[leftmargin=*,itemsep=3pt,topsep=2pt]

    \item \textbf{AV-rich mixture.}
    Internally collected gameplay, human-played Internet gameplay, and general Internet video are selected for high visual and acoustic quality, with speech-rich examples explicitly sampled.
    Reliable camera trajectories are not required, allowing AV-CPT to establish a broad prior over visual appearance, speech, environmental sound, and natural scene dynamics.

\item \textbf{Control-clean mixture.}
    Internally collected gameplay, human-played Internet gameplay, UE simulation, and pose-eligible general Internet video are selected for reliable metric trajectories and smooth, interpretable motion.
    This mixture prioritizes control identifiability and supplies Action-SFT, with the motion-bearing narrative field removed to prevent textual motion leakage.

    \item \textbf{Balanced high-quality mixture.}
    A smaller intersection of reliable trajectory supervision and high-quality audio-visual content combines clear controllable motion with strong visual quality and, when available, environmental sound and speech.
    This mixture supports Joint-FT to consolidate trajectory following with the audio-visual prior.

\end{itemize}

Figure~\ref{fig:control-clean-trajectory-coverage} characterizes the motion
support supplied by the control-clean mixture.
For the coverage analysis, we sample 28,605 trajectories from the control-clean
mixture. Their aggregate distribution spans lateral and forward--backward
displacement, while representative examples include straight, diagonal,
return-loop, and orbiting motion.
This coverage exposes Action-SFT to varied realized camera evolution rather than a narrow set of discrete command templates.

Together, the three mixtures form a data curriculum from audio-visual diversity, through control identifiability, to joint audio-visual and interactive modeling.

\section{Method}
\label{sec:method}

\begin{figure}[ht]
    \centering
    \includegraphics[width=\textwidth]{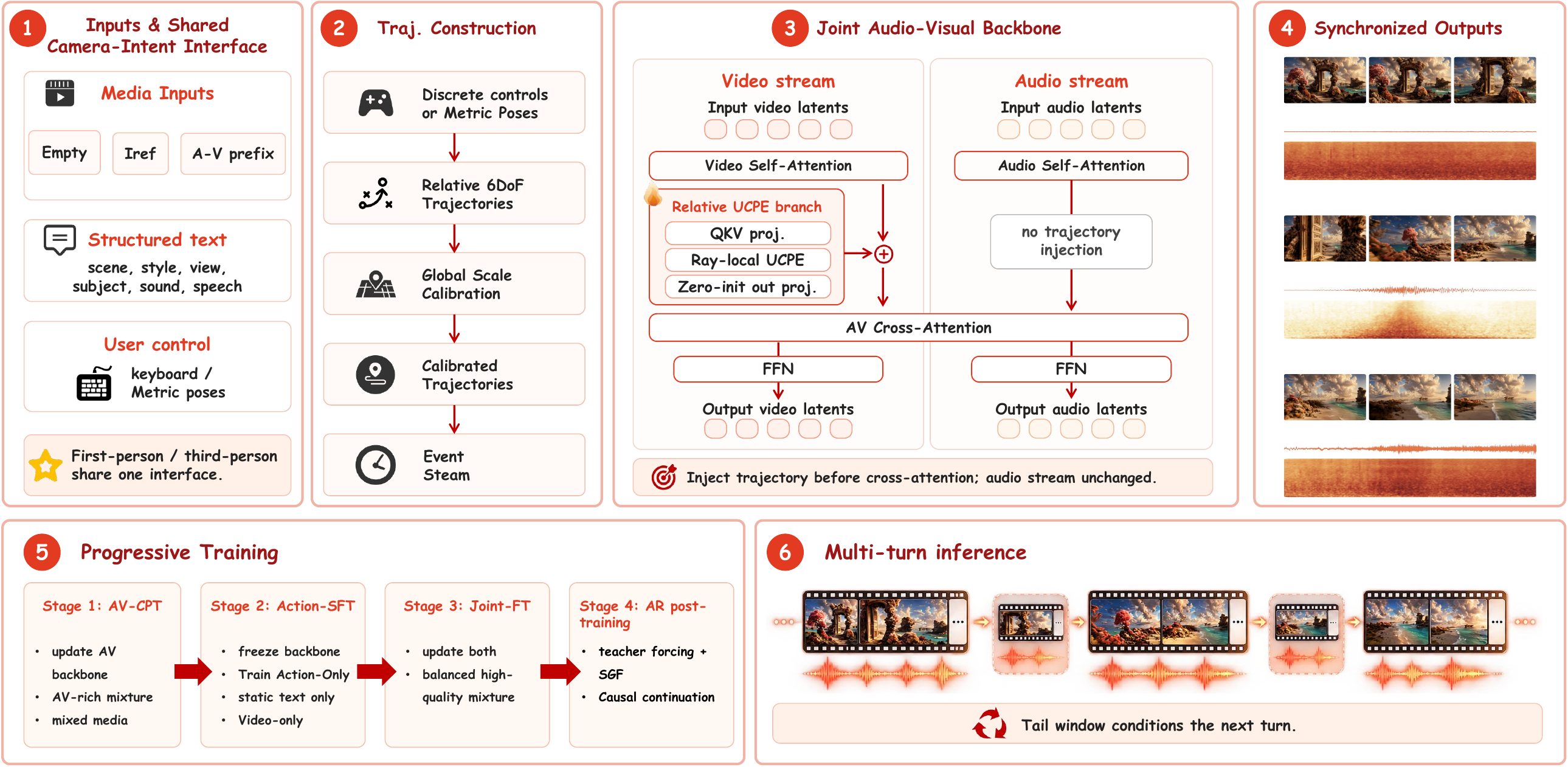}
    \caption{\textbf{Overview of EchoWM.}
    Media context, structured text, and user controls share a unified camera-intent interface.
    Discrete controls or metric poses are converted into globally calibrated relative 6-DoF trajectories and serialized as an event stream.
    A relative UCPE branch injects this trajectory into video self-attention before audio--visual cross-attention, while the audio stream receives no direct trajectory condition.
    Four progressive training stages yield synchronized audio--visual generation and support multi-turn inference through synchronized tail-window conditioning.}
    \label{fig:method-overview}
\end{figure}

\subsection{Overview and Problem Formulation}

Given an optional media context $\mathcal{M}$, a structured text condition
$\mathcal{C}$ following the annotation schema in \cref{sec:data}, and a user
control sequence $U_{1:T}$, our goal is to jointly generate synchronized
video and audio while following the requested camera motion.
The media context may be empty, a reference observation $I_{\mathrm{ref}}$,
or a synchronized audio-visual prefix $(V_{1:\tau},A_{1:\tau})$.

We convert $U_{1:T}$ into a calibrated relative 6-DoF camera trajectory
$P_{1:T}$ and model
\begin{equation}
    p_{\theta,\phi}\!\left(
        V_{1:T},A_{1:T}
        \mid
        \mathcal{M},\mathcal{C},P_{1:T}
    \right),
    \label{eq:world-formulation}
\end{equation}
where $\theta$ denotes the joint audio-visual backbone and $\phi$ denotes
the trajectory-conditioning branch.
The static text fields specify the scene, style, subject, and initial
viewpoint, while $P_{1:T}$ specifies the subsequent camera motion.
Following WorldEvent~\citep{wanteam2026worldevent}, we keep persistent scene
descriptions separate from time-varying motion descriptions and provide
camera evolution explicitly through the geometric trajectory condition.

We build on a pretrained joint audio-visual diffusion transformer and extend
it with metric camera-trajectory conditioning and autoregressive
audio-visual generation.
Heterogeneous user controls and training camera poses are first converted to
a shared relative 6-DoF trajectory with dataset-level translation
calibration.
The resulting trajectory is injected into the video backbone through a
UCPE-based camera branch shared by first- and third-person data.
We train the model progressively, first adapting the audio-visual backbone,
then learning trajectory conditioning with the backbone frozen, and finally
jointly fine-tuning both parameter groups.
Autoregressive audio-visual post-training further adapts the model to causal
generation from its own history.
Figure~\ref{fig:method-overview} summarizes the overall framework.

\subsection{Metric Camera-Intent Conditioning}
\label{sec:camera-conditioning}

We first establish a camera-centric interface that separates user-facing
navigation from the representation consumed by the generator. The resulting
relative trajectory is then calibrated to a shared metric scale across
heterogeneous sources and injected into the video backbone through relative
camera geometry. This decomposition keeps control semantics, translation
amplitude, and neural conditioning explicit while allowing the same pathway to
serve both first- and third-person viewpoints.

\subsubsection{Unified Camera-Intent Interface}
\label{sec:metric-trajectory-interface}

We represent user navigation through the desired camera motion over time rather
than through a system-specific action vocabulary.
In first-person scenes, the trajectory directly describes observer ego-motion.
In third-person scenes, it specifies camera evolution, while the associated
character motion and camera--character coupling are learned from data.
The model receives no explicit character trajectory, controller state, or
camera-rig parameters.
The reference observation and static viewpoint description determine the
initial viewing configuration, while the relative trajectory specifies its
subsequent evolution.

\begin{figure}[ht]
    \centering
    \includegraphics[width=\textwidth]{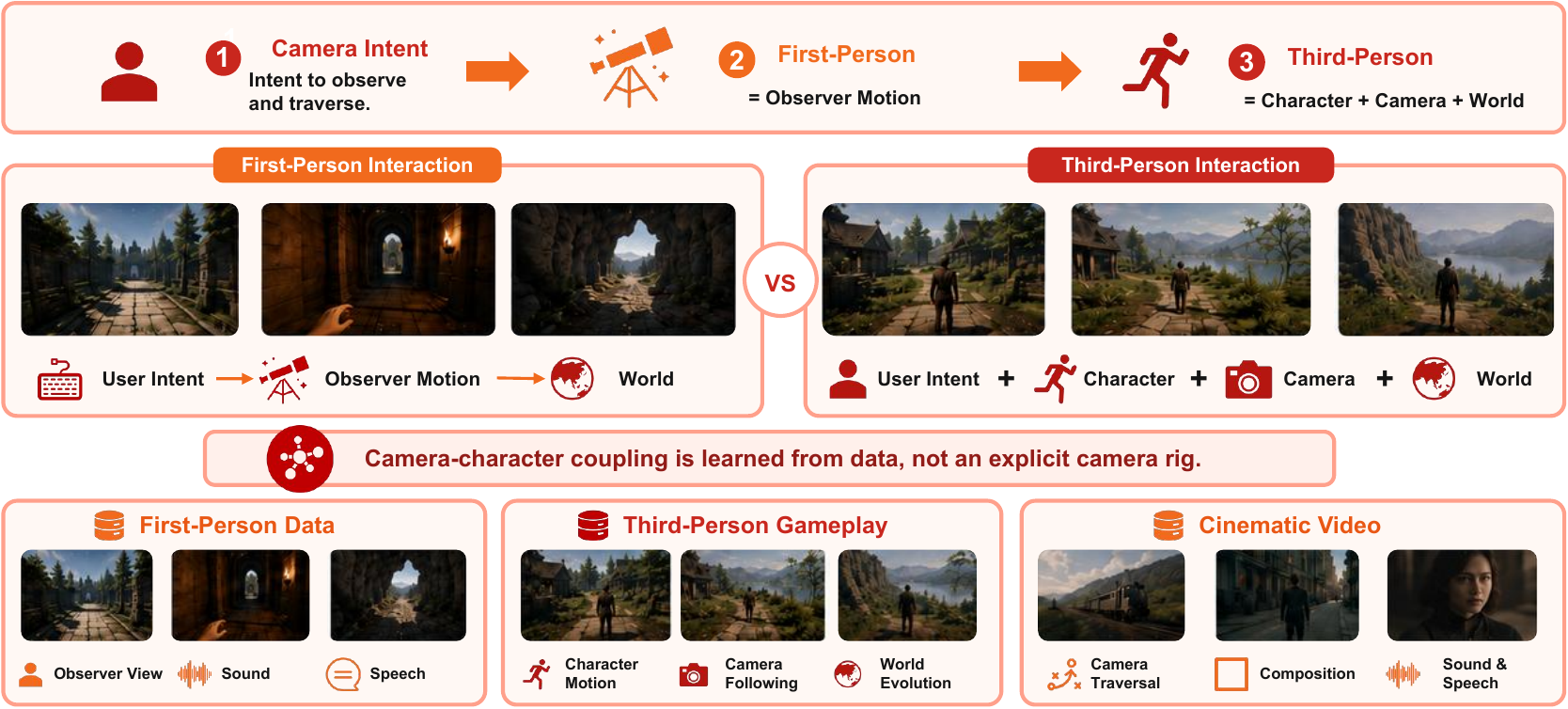}
    \caption{\textbf{Camera intent unifies first- and third-person interaction.}
    The same user intent is realized as observer motion in first-person scenes and as coordinated camera--character--world evolution in third-person scenes. The coupling is learned from heterogeneous data rather than specified by an explicit camera rig.}
    \label{fig:camera-intent}
\end{figure}

Let $T_t\in\mathrm{SE}(3)$ denote the camera-to-world pose at frame $t$.
We express camera motion relative to the initial observation as $\Delta T_t = T_0^{-1}T_t$.
This representation removes dependence on the global rigid coordinate frame
while preserving the continuous magnitude and direction of translation and
rotation.
Although users may interact through discrete navigation commands, we train the
generator on camera trajectories rather than raw action labels for three
reasons:
\begin{itemize}[leftmargin=*,itemsep=2pt,topsep=2pt]

    \item \textbf{Broader data coverage.}
    Frame-aligned action logs are available only for instrumented gameplay and
    simulation, whereas camera trajectories can be recovered from a substantially
    broader range of videos after pose-quality filtering.

    \item \textbf{No controller-dependent inverse mapping.}
    Recovering pseudo-actions from observed camera motion is generally ambiguous:
    the same displacement may result from different user inputs, controller
    dynamics, or camera-smoothing rules.
    This ambiguity is particularly pronounced in third-person scenes, where
    character motion and camera-following behavior jointly determine the observed
    camera trajectory.

    \item \textbf{Continuous 6-DoF motion.}
    Keyboard inputs form a discrete and system-specific control vocabulary.
    In contrast, camera trajectories directly preserve continuous translation
    and rotation, including vertical motion, roll, motion magnitude, and
    translation--rotation coupling.

\end{itemize}

At inference time, users may still interact through discrete navigation
commands.
Let $\mathbf{u}_t$ denote the keyboard state at time $t$.
A fixed mapping $M$, parameterized by the configured per-step translation and
rotation magnitudes, converts the active commands into a local 6-DoF increment
\begin{equation}
    \boldsymbol{\xi}_t
    =
    M\mathbf{u}_t,
    \qquad
    \boldsymbol{\xi}_t
    =
    [\mathbf{v}_t^\top,\boldsymbol{\omega}_t^\top]^\top,
\end{equation}
where $\mathbf{v}_t$ and $\boldsymbol{\omega}_t$ denote translational and
rotational increments in the current camera frame, respectively.
We then integrate these local increments and express the resulting trajectory
relative to its initial pose:
\begin{equation}
    T_t
    =
    T_{t-1}
    \operatorname{Exp}
    \!\left(
        \widehat{\boldsymbol{\xi}}_t
    \right),
    \qquad
    \Delta T_t
    =
    T_0^{-1}T_t,
    \label{eq:action-to-pose}
\end{equation}
where
$\widehat{\boldsymbol{\xi}}_t\in\mathfrak{se}(3)$
is the matrix representation of the 6-DoF increment.
A continuous metric pose sequence bypasses the discrete mapping and
integration, but undergoes the same coordinate conversion and relative
transformation.
Training trajectories recovered in \cref{sec:metric-pose-recovery} and
inference-time navigation commands therefore share the same relative
trajectory representation, which is subsequently calibrated as described in
\cref{sec:pose-normalization}.
This interface separates the user-facing control format from the geometric
condition consumed by the generator, allowing different navigation interfaces
and heterogeneous training sources to share the same trajectory representation.
Its scope is limited to navigation and viewpoint-related controls whose effects
can be represented by camera motion; semantic actor actions such as attacking,
jumping, or object manipulation are not modeled by this interface.

\subsubsection{Global Translation Scale Calibration}
\label{sec:pose-normalization}

Relative poses remove dependence on the global rigid coordinate frame but do
not resolve translation-scale differences across heterogeneous data sources.
For trajectory conditioning, the same numerical displacement should correspond
to a comparable physical camera motion across examples. Per-clip or per-chunk
normalization would erase meaningful displacement differences and can introduce
artificial scale changes at continuation boundaries. Using the largest
displacement in the corpus is instead sensitive to reconstruction outliers.
We therefore estimate one robust scale from the metric training trajectories.

For clip $i$, let
$\Delta T_{i,k}=[\Delta\mathbf{R}_{i,k}\mid\Delta\mathbf{t}_{i,k}]$
denote the relative pose at frame $k$. We compute
\begin{equation}
    m_i=\max_k\left\|\Delta\mathbf{t}_{i,k}\right\|_2,
    \qquad
    s_{\mathrm{global}}
    =Q_{0.9}\!\left(\{m_i\}_{i\in\mathcal{D}_{\mathrm{train}}}\right),
    \label{eq:global-scale}
\end{equation}
where $Q_{0.9}$ denotes the 90th percentile over the training set.
Because trajectory-conditioned training uses fixed-duration clips at a common
temporal sampling rate, $m_i$ provides a comparable measure of clip-level
translation extent across sources.
We retain trajectories whose maximum displacement does not exceed
$s_{\mathrm{global}}$ rather than clipping larger trajectories, which would
alter their displacement magnitude. For each retained trajectory,
\begin{equation}
    \widehat{\mathbf{t}}_{i,k}
    =\frac{\Delta\mathbf{t}_{i,k}}{s_{\mathrm{global}}},
    \qquad
    P_{i,k}=\left[\Delta\mathbf{R}_{i,k}\mid\widehat{\mathbf{t}}_{i,k}\right].
    \label{eq:translation-normalization}
\end{equation}
Rotations remain unchanged, and the same $s_{\mathrm{global}}$ is used for all
training and inference trajectories. Together with the common temporal
sampling rate, this preserves displacement differences across examples and
avoids scale-induced changes in motion speed at continuation boundaries.

\subsubsection{Relative Trajectory Conditioning}
\label{sec:camera-interface}

Given the calibrated trajectory $P_{1:T}$, we next inject camera geometry
into the video backbone.
In our comparison setup, Pl\"ucker-based conditioning represents the
world-space ray associated with spatial cell $s$ at frame $t$ as
\begin{equation}
    \mathcal{P}_{t,s}
    =
    \left(
        \mathbf{d}_{t,s},
        \mathbf{m}_{t,s}
    \right),
    \qquad
    \mathbf{m}_{t,s}
    =
    \mathbf{o}_t\times\mathbf{d}_{t,s},
    \label{eq:plucker-ray}
\end{equation}
where $\mathbf{d}_{t,s}$ is the ray direction and $\mathbf{o}_t$ is the
camera center.
Because the moment term explicitly contains $\mathbf{o}_t$, this absolute
representation changes with the chosen world origin.
The baseline additionally materializes dense ray features and uses a
separate camera encoder to learn relations across frames.
We instead use Unified Camera Positional Encoding
(UCPE)~\citep{zhang2025ucpe}, which introduces relative ray geometry
directly inside attention.
For a global rigid transformation $G$, $(GT_i)^{-1}(GT_j)=T_i^{-1}T_j$, so pairwise camera geometry is unchanged by a global rigid change of coordinates. Translation scale is handled separately by the calibration in~\cref{sec:pose-normalization}.
Let $i=(t,s)$ index a video latent token at frame $t$ and spatial cell $s$,
with homogeneous image coordinate $\mathbf{p}_s$.
We interpret the calibrated pose
$P_t=[\mathbf{R}_t\mid\mathbf{o}_t]$ as a camera-to-reference transform,
with the first camera defining the reference frame.
Given this pose and intrinsics $K_t$, the corresponding reference-space ray
direction is
\begin{equation}
    \mathbf{d}_{t,s}
    =
    \mathbf{R}_t
    \operatorname{norm}
    \!\left(
        K_t^{-1}\mathbf{p}_s
    \right).
    \label{eq:ucpe-ray-direction}
\end{equation}
Following UCPE, we construct a local ray frame using this direction and the
camera image axes.
Let $\mathbf{v}^{\mathrm{down}}_t$ denote the image-down direction:
\begin{equation}
\begin{aligned}
    \mathbf{z}_{t,s}
        &= \mathbf{d}_{t,s},\\
    \mathbf{x}_{t,s}
        &=
        \operatorname{norm}
        \!\left(
            \mathbf{v}^{\mathrm{down}}_t
            \times
            \mathbf{z}_{t,s}
        \right),\\
    \mathbf{y}_{t,s}
        &=
        \mathbf{z}_{t,s}
        \times
        \mathbf{x}_{t,s}.
\end{aligned}
    \label{eq:ucpe-ray-frame}
\end{equation}
Together with camera center $\mathbf{o}_t$, this basis defines the
ray-to-reference transform $\mathbf{T}^{\mathrm{wr}}_i$.
Its inverse
$\mathbf{D}_i=(\mathbf{T}^{\mathrm{wr}}_i)^{-1}$
maps reference coordinates into the local ray frame.
We inject this geometry through a lightweight camera-attention branch
parallel to the original video self-attention.
For a head of width $d$, half of the channels use the tiled ray transform
$\mathbf{G}_i=\mathbf{I}_{d/8}\otimes\mathbf{D}_i$, while the remaining
channels retain spatiotemporal RoPE.
With branch-specific QKV projections, the camera attention is
\begin{equation}
\begin{aligned}
    \widetilde{\mathbf{q}}^{c}_i
        &=
        (\mathbf{G}_i^\top\oplus\operatorname{RoPE}_i)
        \mathbf{q}^{c}_i,\\
    (\widetilde{\mathbf{k}}^{c}_i,\widetilde{\mathbf{v}}^{c}_i)
        &=
        (\mathbf{G}_i^{-1}\oplus\operatorname{RoPE}_i)
        (\mathbf{k}^{c}_i,\mathbf{v}^{c}_i),\\
    \mathbf{o}^{c}_i
        &=
        (\mathbf{G}_i\oplus\operatorname{RoPE}_i^{-1})
        \operatorname{Attn}_{\mathrm{cam}}
        \!\left(
            \widetilde{\mathbf{q}}^{c},
            \widetilde{\mathbf{k}}^{c},
            \widetilde{\mathbf{v}}^{c}
        \right)_i .
\end{aligned}
    \label{eq:ucpe-attention}
\end{equation}
The resulting query--key interaction depends on the relative ray transform
$\mathbf{G}_i\mathbf{G}_j^{-1}$ between tokens $i$ and $j$.
The camera-branch output is merged into the original video attention through
a zero-initialized projection.
The branch-specific QKV and output projections constitute $\phi$.
Zero initialization preserves the pretrained video pathway at the start of
trajectory fine-tuning and lets the branch learn only the residual required
for camera conditioning.
The same branch is shared across first- and third-person data.
The static viewpoint field specifies the initial observer--subject
relationship, while $P_{1:T}$ specifies subsequent camera motion.
No view-specific network or discrete action embedding is introduced.
The camera branch operates only on video tokens; audio tokens receive no
direct trajectory conditioning.
We use the relative-ray component of UCPE and omit its optional Lat-Up
encoding, since the trajectory is defined relative to the initial camera frame.

\subsection{Progressive Audio-Visual Control Training}
\label{sec:training-framework}

\begin{figure}[t!]
    \centering
    \includegraphics[width=0.92\textwidth]{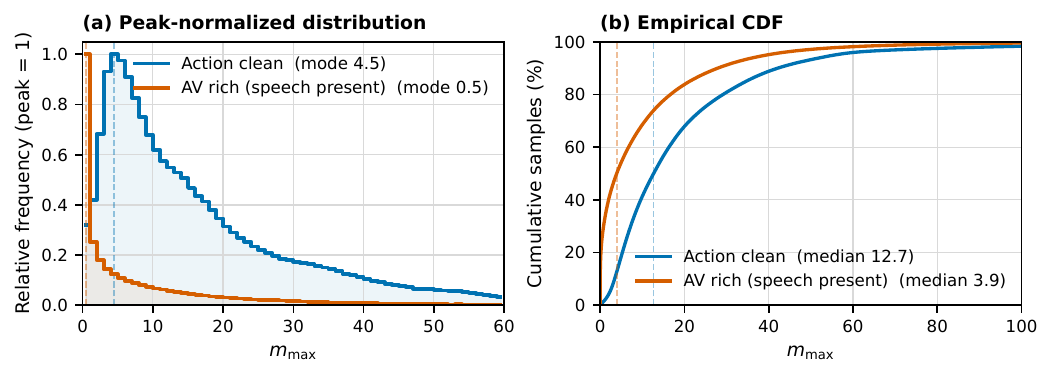}
    \caption{\textbf{Pose-magnitude distributions for global scale calibration.}
    The plots show the peak-normalized distribution and empirical CDF of the per-clip maximum translation magnitude $m_i$ for the action-clean and AV-rich speech-present subsets.
    Action-clean clips contain broader motion magnitudes (mode $4.5$, median $12.7$), whereas speech-present AV-rich clips concentrate at smaller motions (mode $0.5$, median $3.9$).
    The heterogeneous support motivates one robust dataset-level percentile scale rather than independently normalizing each source.}
    \label{fig:pose-distribution}
\end{figure}

Reliable audio-visual supervision and reliable trajectory supervision are
concentrated in different parts of the training data.
As shown in \cref{fig:pose-distribution}, AV-rich examples provide stronger
audio supervision but typically cover a narrower range of camera motion,
whereas control-clean examples contain more reliable and diverse
trajectories.
We therefore separate training into three stages.
AV-CPT first adapts the joint audio-visual backbone, Action-SFT then learns
the trajectory-conditioning branch with the backbone frozen, and Joint-FT
finally updates both parameter groups on the balanced high-quality subset.

\noindent{\textbf{Stage 1: Audio-Visual Continued Pretraining.}}
The pretrained model already supports joint audio-visual generation, but the
interactive training data differ in visual appearance, motion statistics,
speech, ambience, and interaction-related sound.
We first adapt the complete backbone $\theta$ on the AV-rich mixture without
trajectory conditioning.
The full structured annotation is retained, including narrative, speech,
and sound.
For each example, the clean media context $\mathcal{M}$ is sampled from an
empty condition, a reference frame $I_{\mathrm{ref}}$, or a synchronized
audio-visual prefix $(V_{1:\tau},A_{1:\tau})$.
These cases train text-conditioned generation, reference-frame-conditioned
generation, and synchronized continuation, respectively.
When a reference frame or prefix is provided, the exposed tokens remain
clean and are excluded from the denoising loss.
We optimize
\begin{equation}
\begin{aligned}
    \theta_{\mathrm{AV}}
    =
    \arg\min_{\theta}\;
    \mathbb{E}_{
        (V,A,\mathcal{C}_{\mathrm{AV}})
        \sim\mathcal{D}_{\mathrm{AV}},\,
        \mathcal{M}
        \sim q_{\mathrm{cond}}(\cdot\mid V,A)
    }
    \Big[
        &
        \mathcal{L}_{\mathrm{V}}
        \!\left(
            V
            \mid
            \mathcal{C}_{\mathrm{AV}},
            \mathcal{M};
            \theta
        \right)
        \\
        &+
        \lambda_{\mathrm{A}}
        \mathcal{L}_{\mathrm{A}}
        \!\left(
            A
            \mid
            \mathcal{C}_{\mathrm{AV}},
            \mathcal{M};
            \theta
        \right)
    \Big],
\end{aligned}
    \label{eq:av-cpt-objective}
\end{equation}
where both losses are evaluated only on regions selected for generation by
the sampled media condition.

\noindent{\textbf{Stage 2: Action Fine-Tuning.}}
We refer to this stage as Action-SFT because it trains responsiveness to the
user-facing navigation interface, although supervision is provided by camera
trajectories rather than raw keyboard labels.
Starting from $\theta_{\mathrm{AV}}$, we freeze the audio-visual backbone and
train only the camera-branch QKV projections and zero-initialized output
projection, collectively denoted by $\phi$.
Training uses the control-clean mixture, which favors reliable metric
trajectories and visually unambiguous camera motion.
To prevent text from directly specifying the target motion, we retain only
the static annotation fields
$\mathcal{C}_{\mathrm{static}}
=
\{c_{\mathrm{scene}},c_{\mathrm{style}},c_{\mathrm{view}},
c_{\mathrm{subject}}\}$.
The narrative field and other motion-dependent descriptions are removed, so
$P_{1:T}$ is the only condition that specifies the target camera evolution.
We disable the audio loss and sample
$\mathcal{M}_{\mathrm{SFT}}\in\{\varnothing,I_{\mathrm{ref}}\}$,
with reference-frame-conditioned examples sampled more frequently than
unconditional ones.
The reference frame constrains scene appearance and the initial viewpoint,
leaving the trajectory to specify subsequent camera motion.
We optimize
\begin{equation}
    \phi_{\mathrm{SFT}}
    =
    \arg\min_{\phi}\;
    \mathbb{E}_{
        \mathcal{M}_{\mathrm{SFT}}
        \sim q_{\mathrm{SFT}}
    }
    \left[
        \mathcal{L}_{\mathrm{V}}
        \!\left(
            V
            \mid
            \mathcal{M}_{\mathrm{SFT}},
            \mathcal{C}_{\mathrm{static}},
            P_{1:T};
            \theta_{\mathrm{AV}},
            \phi
        \right)
    \right].
    \label{eq:action-sft-objective}
\end{equation}
The trajectory is used only as a conditioning signal and is not itself a
prediction target.
No explicit action-to-audio objective is introduced in this stage.

\noindent{\textbf{Stage 3: Joint Fine-Tuning.}}
AV-CPT and Action-SFT optimize different parameter subsets on different data
mixtures.
We therefore perform a final Joint-FT stage on the smaller balanced
high-quality subset, for which both camera trajectories and audio-visual
signals are reliable.

The text condition retains the static fields together with speech and sound,
while the narrative field remains excluded.
Starting from
$(\theta_{\mathrm{AV}},\phi_{\mathrm{SFT}})$,
we update both parameter groups with a reduced learning rate:
\begin{equation}
    (\theta^{\star},\phi^{\star})
    =
    \arg\min_{\theta,\phi}\;
    \mathbb{E}
    \left[
        \mathcal{L}_{\mathrm{V}}
        (\theta,\phi;P_{1:T})
        +
        \lambda_{\mathrm{A}}
        \mathcal{L}_{\mathrm{A}}
        (\theta,\phi)
    \right].
    \label{eq:joint-ft-objective}
\end{equation}
The camera branch remains restricted to the video stream.
The reduced learning rate allows the backbone and trajectory branch to be
updated jointly while limiting disruption to the audio-visual and control
capabilities learned in the preceding stages.

\subsection{Streaming Audio-Visual Post-Training}
\label{sec:post-training}

We progressively adapt a pretrained bidirectional, multi-step audio-visual diffusion model into a high-throughput, low-latency streaming autoregressive generator. This transition requires two key transformations: converting bidirectional temporal attention into causal autoregressive computation for online generation of synchronized audio-video chunks, and compressing iterative denoising into a few-step sampler that remains stable under self-generated histories. To achieve this, we first apply audio-visual teacher forcing~\cite{williams1989learning,chen2024diffusion} to initialize a causal streaming generator from the bidirectional backbone. We then introduce short-horizon Self-Gradient Forcing (SGF)~\cite{huang2026self,zhuang2026sgf} to train on self-generated rollout states while distilling few-step generation, and further extend SGF to long-horizon self-rollouts to improve robustness and stability during sustained streaming inference.
All autoregressive post-training stages retain the trajectory-conditioning
branch and jointly update the generator parameters $(\theta,\phi)$ unless
otherwise stated.

\begin{figure}[ht]
    \centering
    \includegraphics[width=0.82\linewidth]{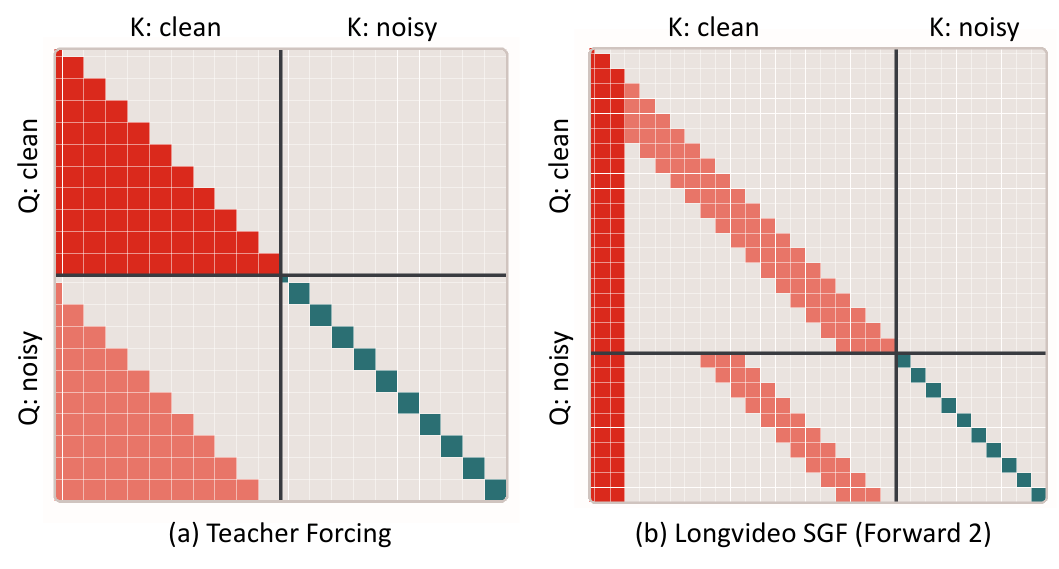}
    \caption{
    Chunk-level causal attention patterns used for streaming post-training.
    (a) The audio-visual teacher-forcing pattern used by short-horizon
    reconstruction: each noisy query attends to the preceding clean context and
    the noisy tokens of its current chunk, while clean targets and future chunks
    are excluded.
    (b) The mask used by Forward 2 of long-horizon SGF, where access to clean
    history is additionally constrained by the sink-plus-FIFO KV-cache policy.
    }
    \label{fig:causal-mask}
\end{figure}

\subsubsection{Audio-Visual Teacher Forcing}
\label{sec:causal-av-teacher-forcing}

Let $m\in\{v,a\}$ index the modality, with $v$ denoting video and $a$ denoting
audio. We divide both latent streams into $N$ temporally aligned macro-chunks.
The clean latent of modality $m$ in chunk $i$ is denoted by $x_i^m$, where
$i\in\{1,\ldots,N\}$. An aligned video-audio pair spans the same time interval,
although the two chunks may contain different numbers of tokens. For each pair,
we sample a shared noise level $\sigma_i\in(0,1)$ and construct
\begin{equation}
  x_{\sigma_i,i}^{m}
  =(1-\sigma_i)x_i^{m}+\sigma_i\epsilon_i^{m},
  \qquad \epsilon_i^{m}\sim\mathcal{N}(0,I),
  \label{eq:ar-block-noising}
\end{equation}
where $\epsilon_i^m$ is Gaussian noise and $I$ is the identity covariance
matrix. The shared $\sigma_i$ places the paired audio and video chunks at a
consistent diffusion stage.

To enable causal generation without sacrificing training parallelism, we
concatenate the clean latents with their noisy counterparts. Let
$\mathcal{B}_{i,\mathrm{clean}}$ and $\mathcal{B}_{i,\mathrm{noisy}}$ denote
the clean and noisy latent blocks of the $i$-th paired audio-visual chunk,
respectively. We impose the following causal attention pattern:
\begin{equation}
  \mathcal{S}(\mathcal{B}_{i,\mathrm{clean}})
  =\{\mathcal{B}_{j,\mathrm{clean}}:j\leq i\},
  \qquad
  \mathcal{S}(\mathcal{B}_{i,\mathrm{noisy}})
  =\{\mathcal{B}_{j,\mathrm{clean}}:j<i\}
   \cup\{\mathcal{B}_{i,\mathrm{noisy}}\},
  \label{eq:ar-teacher-forcing-mask}
\end{equation}
where $\mathcal{S}(\cdot)$ denotes the set of latent blocks accessible to the
corresponding attention query. Under this mask, each noisy chunk can attend to
all preceding clean audio-visual chunks and to the noisy tokens within the
current chunk, while the corresponding clean target and all future chunks are
masked out. We apply the same causal constraint to video self-attention, audio
self-attention, and both directions of audio-video cross-attention. As
illustrated in Fig.~\ref{fig:causal-mask}(a), this causal attention pattern
enables each chunk to use the complete preceding clean history while preventing
information leakage from the current clean target and future chunks. This
converts the original bidirectional temporal attention into chunk-wise causal
attention, while retaining parallel prediction of all chunks during training.

We additionally condition the model on a global condition $\mathcal{C}$ and a
temporally aligned trajectory sequence $P^{(1:N)}$, where $P^{(i)}$ denotes the
trajectory segment associated with chunk $i$. Consistent with the
autoregressive generation process, chunk $i$ only uses the trajectory available
up to the current chunk, denoted by $P^{(\leq i)}$.

Given the causally masked sequence, the model jointly predicts the flow
velocities of the audio and video latents in a single forward pass. We retain
the standard flow-matching objective of the pretrained model:
\begin{equation}
  \mathcal{L}_{\mathrm{TF}}
  =
  \sum_{m\in\{v,a\}}\lambda_m
  \mathbb{E}
  \left[
    \sum_{i=1}^{N}
    \left\|
      v_\theta^m
      (x_{\sigma_i,i}^{m},\sigma_i
      \mid \mathcal{C},P^{(\leq i)})
      -
      (\epsilon_i^m-x_i^m)
    \right\|_2^2
  \right],
  \label{eq:ar-teacher-forcing-loss}
\end{equation}
where $v_\theta^m$ denotes the predicted flow velocity for modality $m$,
$\lambda_m$ balances the video and audio losses, and $\theta$ denotes the
generator parameters. Thus, audio-visual teacher forcing adapts the pretrained
bidirectional diffusion model to causal, chunk-wise autoregressive generation
while preserving its original flow-matching training objective.

\subsubsection{Short-Horizon Audio-Visual Self-Gradient Forcing}
\label{sec:short-sgf}

The causal teacher-forcing model provides an effective initialization for
autoregressive generation, but two discrepancies remain within the temporal
horizon of the base model. First, teacher-forcing training conditions the
generator on ground-truth histories, whereas inference relies on its own
generated outputs. Second, the original diffusion model requires multiple
denoising steps for each audio-visual chunk. We address these discrepancies
using Self-Gradient Forcing (SGF)~\citep{zhuang2026sgf}, which trains on
self-generated autoregressive histories while recovering gradients through
their reconstructed causal representations. We additionally apply
Distribution Matching Distillation (DMD)~\citep{yin2024dmd} to convert the
original multi-step generator into a few-step sampler.

\noindent{\textbf{Forward 1: autoregressive self-rollout.}}
We first run a complete chunk-wise rollout without gradient tracking. This
rollout uses the same few-step sampler, causal attention configuration, and
trajectory conditioning as autoregressive inference. When generating chunk
$i$, the model is conditioned on the currently available trajectory
$P^{(\leq i)}$. For instance, with a four-step sampler, the model performs four
denoising steps to produce each synchronized audio-visual chunk and then
executes an additional clean-context forward pass at $t=0$ to populate the
causal KV cache used by subsequent chunks.

An exit step $t_e$ is sampled during the rollout. For every chunk $i$, we
record the noisy audio-visual state at this step,
$r_i=(r_i^v,r_i^a)$, together with the final generated clean latents
$\widehat{x}_i=(\widehat{x}_i^v,\widehat{x}_i^a)$. All operations in this
forward pass are performed under \texttt{no\_grad}. Forward 1 therefore
captures the autoregressive states encountered during few-step inference
without retaining the computation graph of the sequential sampling process.

\noindent{\textbf{Forward 2: audio-visual context-gradient reconstruction.}}
We subsequently replay the recorded exit-step computation in a differentiable
parallel forward pass. The generated clean latents
$\{\widehat{x}_i\}_{i=1}^{N}$ are treated as stop-gradient inputs and evaluated
at the clean context timestep $t=0$, while the recorded noisy states
$\{r_i\}_{i=1}^{N}$ retain their sampled timestep $t_e$. Clean-context tokens
and noisy target tokens are then arranged using the same causal construction
as Eq.~\eqref{eq:ar-teacher-forcing-mask}.

The resulting attention pattern is shown in
Fig.~\ref{fig:causal-mask}(a). A noisy query associated with chunk $i$ can
attend to the clean history from preceding chunks and to the noisy tokens of
the current chunk. Clean targets, future clean chunks, and future noisy chunks
remain inaccessible. Clean-context queries are likewise restricted to causal
clean history. This mask reproduces the context--target dependencies of the
serial rollout while allowing the exit-step predictions for all chunks to be
reconstructed in parallel.

In our LTX-2.3-based audio-visual generator, this chunk-level causal mask is
applied to all five attention operations that involve temporally indexed
rollout information: video self-attention, audio self-attention,
audio-to-video (A2V) attention, video-to-audio (V2A) attention, and UCPE
attention. The video and audio self-attention masks enforce causality within
each modality, while the A2V and V2A masks impose the same temporal constraint
on cross-modal information exchange. The UCPE attention follows the
corresponding chunk-level visibility pattern for trajectory-conditioned
features. Text cross-attention is unchanged because the global text condition
is available throughout the complete rollout.

Although the clean and noisy latents recorded in Forward 1 are detached, the
generator recomputes their context representations with gradient tracking in
Forward 2. In particular, the clean-context forward reconstructs the
attention projections and K/V representations through which generated
audio-visual chunks are encoded as causal history. A loss on a later chunk can
therefore propagate through its causal attention operations into the
reconstructed K/V representations of preceding generated chunks. This
restores supervision for the audio-visual context-writing computation while
avoiding backpropagation through the denoising decisions and serial cache
updates of Forward 1. The reconstructed clean predictions are denoted by
$\{\bar{x}_i\}_{i=1}^{N}$, where
$\bar{x}_i=(\bar{x}_i^v,\bar{x}_i^a)$.

\noindent{\textbf{Distribution-matching objective.}}
We apply DMD to the recovered predictions under the self-generated
autoregressive histories. Unlike teacher-forcing supervision, these
predictions are conditioned on the generator's own preceding audio-visual
outputs. The resulting objective therefore combines few-step diffusion
distillation with adaptation to the history distribution encountered during
autoregressive inference.

For a sampled diffusion timestep $s$, the recovered clean audio-visual latent
is perturbed as
\begin{equation}
    \bar{x}_{s,i}
    =
    (1-s)\bar{x}_i+s\eta_i,
    \qquad
    \eta_i\sim\mathcal{N}(0,I),
    \label{eq:sgf-score-noising}
\end{equation}
where $\eta_i$ is independently sampled Gaussian noise. The generator is
initialized from the causal model obtained after teacher forcing. Both the
real-score model $s_{\mathrm{real}}$ and the fake-score model
$s_{\mathrm{fake}}$ are initialized from the pretrained bidirectional
diffusion model. The real-score model remains frozen and represents the
reference distribution, whereas the fake-score model is trainable and tracks
the evolving distribution of the few-step generator.

The generator is optimized using
\begin{equation}
    \mathcal{L}_{\mathrm{DMD}}
    =
    \mathbb{E}_{i,s,\eta_i}
    \left[
        w(s)
        \left\langle
        \operatorname{sg}\!\left(
            s_{\mathrm{fake}}(\bar{x}_{s,i},s)
            -
            s_{\mathrm{real}}(\bar{x}_{s,i},s)
        \right),
        \bar{x}_i
        \right\rangle
    \right],
    \label{eq:sgf-dmd-loss}
\end{equation}
where $\operatorname{sg}(\cdot)$ denotes stop-gradient and $w(s)$ is a
diffusion-dependent weighting factor. For clarity, the global condition
$\mathcal{C}$ and trajectory condition $P^{(\leq i)}$ are omitted from the
score-model notation. The score difference directs the few-step generator
toward the distribution represented by the pretrained multi-step model, while
the dependence of $\bar{x}_i$ on the differentiable reconstruction preserves
gradients through the causal context representations.

\noindent{\textbf{Fake-score updates.}}
Following each generator update, we perform multiple updates of the
fake-score model. Generated samples are detached and used to optimize
$s_{\mathrm{fake}}$ with the standard flow-matching objective, enabling it to
follow the changing generator distribution.

Short-horizon audio-visual SGF thus exposes the generator to its own
autoregressive histories, restores supervision through the reconstructed
causal context, and establishes few-step generation. The resulting model
serves as the initialization for long-horizon streaming training.

\subsubsection{Long-Horizon Audio-Visual Self-Gradient Forcing}
\label{sec:long-video-sgf}

Short-horizon SGF adapts the generator to self-generated states within the
temporal range of the base model. During sustained streaming generation,
however, the model must operate on histories containing errors accumulated
over a much longer sequence. We therefore continue self-rollout training on
substantially longer synchronized audio-visual trajectories. The few-step
sampler and DMD objective are retained from the short-horizon stage, while the
extended rollout exposes the model to progressively shifted autoregressive
states.

\noindent{\textbf{Long-horizon rollout.}}
A long-horizon trajectory is formed by sequentially generating multiple
temporal segments, each using the rollout length adopted during short-horizon
SGF. To maintain continuity between adjacent segments, the terminal video
frame of segment $k$ is used as the initial frame of segment $k+1$. The model
does not restart from an independent visual state at each boundary. Instead,
later segments are generated from the preceding self-generated trajectory and
therefore inherit the prediction errors accumulated in earlier segments. This
provides training exposure to the states encountered during sustained
world-model rollout.

\noindent{\textbf{Bounded causal context.}}
A full causal KV cache would grow continuously with the rollout duration. We
instead maintain a sink-plus-FIFO cache for each modality. Let $S_m$ denote the
number of persistent sink chunks and $R_m$ the number of recent chunks retained
for modality $m$. When generating chunk $i$, the available historical context
is
\begin{equation}
  \mathcal{H}_i^m
  =
  \underbrace{
    \bigl(
      \widehat{x}_1^m,\ldots,
      \widehat{x}_{\min(S_m,i-1)}^m
    \bigr)
  }_{\text{attention sink}}
  \mathbin{\Vert}
  \underbrace{
    \bigl(
      \widehat{x}_{\max(S_m+1,i-R_m)}^m,\ldots,
      \widehat{x}_{i-1}^m
    \bigr)
  }_{\text{recent FIFO history}},
  \label{eq:sink-sliding-window}
\end{equation}
where $\mathbin{\Vert}$ denotes ordered concatenation. The sink retains a
fixed prefix of the generated trajectory, while the FIFO component preserves
the most recent context. Consequently, the number of historical chunks
directly available at each attention layer is bounded by $S_m+R_m$, regardless
of the total rollout duration.

Forward 2 reconstructs the long trajectory using the attention pattern shown
in Fig.~\ref{fig:causal-mask}(b). Relative to the short-horizon mask in
Fig.~\ref{fig:causal-mask}(a), the clean causal history available to each
query is further restricted by the sink-plus-FIFO policy. A noisy query
attends to the retained clean sink and FIFO context, together with the noisy
tokens of its current chunk. The same visibility pattern is used during
autoregressive inference and during the differentiable reconstruction.

\noindent{\textbf{Layer-wise long-range context gradients.}}
The sink-plus-FIFO window limits the K/V entries that a query accesses
\emph{directly at a single Transformer layer}, but it does not confine the
effective context gradient to only those entries. A retained FIFO
representation at an upper layer has already incorporated information from
the causal context visible to that chunk in preceding layers. During
backpropagation, a loss on a later chunk first reaches the sink and FIFO K/V
representations visible at the current layer. The gradient can then continue
through the lower-layer computations that produced these representations and
reach earlier K/V representations through intermediate chunks. Stacking the
causal attention layers therefore produces a multi-layer gradient path whose
effective temporal reach can be longer than the direct sink-plus-FIFO window
of any individual layer.

This layer-wise propagation does not introduce backpropagation through the
serial rollout. The latents, denoising states, and KV-cache trajectory
recorded in Forward 1 remain detached. Gradients are propagated only through
the attention and context representations recomputed in Forward 2. Thus, the
cache policy keeps the direct attention fan-in bounded, while the
reconstructed multi-layer computation allows later losses to provide
supervision beyond the K/V entries directly visible at the current layer.

\noindent{\textbf{Score-model evaluation range.}}
The generator retains its original RoPE configuration throughout
long-horizon rollout, independently of the sink-plus-FIFO cache policy.
Because the real-score and fake-score models preserve the bidirectional
architecture of the pretrained fixed-length diffusion model, their DMD
evaluations are restricted to the temporal range observed during pretraining.

\noindent{\textbf{Long-horizon DMD optimization.}}
We apply DMD supervision to every rollout segment rather than only to the
final part of the trajectory. Let
$\mathcal{L}_{\mathrm{DMD}}^{(k)}$ denote the DMD loss computed from the
recovered audio-visual predictions of segment $k$. For a trajectory containing
$K$ rollout segments, the training objective is
\begin{equation}
  \mathcal{L}_{\mathrm{Long\text{-}SGF}}
  =
  \sum_{k=1}^{K}
  \mathcal{L}_{\mathrm{DMD}}^{(k)} .
  \label{eq:long-sgf-loss}
\end{equation}
Each segment is consequently supervised at a different stage of the
self-generated trajectory. Losses on later segments are evaluated under
histories that already reflect the errors accumulated in preceding segments.
Together with the reconstructed multi-layer context gradients, this training
adapts the few-step generator to the progressively changing state distribution
encountered during long-horizon world-model rollout.

\subsection{Interactive Inference}
\label{sec:continuation}
\noindent\textbf{Bidirectional multi-turn continuation.}
At inference time, multi-turn generation reuses synchronized video and audio
from the preceding segment as clean context for the next turn.
For turn $j>1$, we extract
\begin{equation}
    \mathcal{M}^{(j-1)}
    =
    \operatorname{Tail}_{w}
    \!\left(
        V^{(j-1)},A^{(j-1)}
    \right),
    \label{eq:multi-turn-context}
\end{equation}
and generate
\begin{equation}
    \left(
        V^{(j)},A^{(j)}
    \right)
    \sim
    p_{\theta^{\mathrm{PT}},\phi^{\mathrm{PT}}}
    \!\left(
        \cdot
        \mid
        \mathcal{M}^{(j-1)},
        \mathcal{C}^{(j)},
        P_{1:T}^{(j)}
    \right).
    \label{eq:multi-turn-inference}
\end{equation}
Each turn may specify a different navigation trajectory.
Because the context window contains synchronized video and audio, visual and
acoustic continuation are conditioned on the same recent history.
The generated segment then provides the tail context for the following turn.
This tail-reencoding procedure is the bidirectional continuation mode. We use
a separate persistent-state procedure for causal streaming inference, described
next, so the two modes differ in how they manage historical context rather than
in the trajectory interface or model parameterization.


\noindent\textbf{Causal streaming inference.}
We keep a persistent streaming KV cache instead of restarting from a decoded
tail. At inference, we maintain separate streaming KV caches for video and
audio, along
with the camera-attention KV states associated with video tokens, each following
the same sink-plus-local structure used during
training. The sink retains a persistent long-term prefix, while the local
window stores the most recent context. As old local tokens are removed, we
rebase temporal RoPE and camera encodings over the retained cache to maintain
a consistent positional layout.

Let $\mathcal{K}^{(t)}$ denote the retained cache state before the next block.
Given text condition $\mathcal{C}$ and trajectory $P$, we sample
\begin{equation}
    (V,A)
    \sim
    p_{\theta^{\mathrm{PT}},\phi^{\mathrm{PT}}}
    \bigl(\cdot \mid \mathcal{K}^{(t)}, \mathcal{C}, P\bigr).
    \label{eq:inference-memory}
\end{equation}
The persistent sink provides long-term context throughout the autoregressive
rollout without requiring an external memory bank.

\section{Evaluation}
\label{sec:experiments}

We evaluate \model on two public benchmarks for direct comparison
with recent world models.
WBench Navigation measures broad interactive-world performance across visual
quality, setting adherence, interaction, consistency, and physical plausibility,
whereas SANA-WM-Bench provides a more focused evaluation of short- and
long-horizon camera-trajectory following and visual persistence.
Together, these benchmarks provide complementary evidence for generation
quality, controllability, and long-horizon world consistency.
We further supplement the quantitative results with qualitative analysis,
human evaluation, and ablation studies.

\subsection{WBench Navigation}
\label{sec:navigation-eval}

\begin{table}[ht]
    \centering
    \small
    \caption{
    Comparison on the 158-case WBench Navigation split~\citep{ying2026wbench}.
    Results are sorted by Average.
    }
    \label{tab:wbench-navigation}
    \resizebox{1.0\textwidth}{!}{%
    \begin{tabular}{l c c c c c c}
        \toprule
        Model
        & Avg.
        & Quality
        & Setting
        & Interaction
        & Consistency
        & Physical \\
        \midrule

        \textcolor{EchoAccent}{\textbf{\model}}
        & \textbf{81.7}
        & 81.5
        & 79.4
        & \underline{87.2}
        & \underline{89.8}
        & 70.6 \\

        \textcolor{EchoAccent}{\textbf{\model{}-Flash}}
        & \underline{81.0}
        & 81.1
        & 88.3
        & \textbf{87.9}
        & 77.5
        & 70.1 \\

        \midrule

        HiDream-O1-World~\citep{ying2026wbench}
        & 80.9
        & 81.0
        & 82.2
        & 80.0
        & 88.0
        & \textbf{73.3} \\

        Alaya-EVOKE~\citep{ying2026wbench}
        & 80.8
        & \textbf{82.8}
        & 83.8
        & 78.6
        & 86.9
        & \underline{72.1} \\

        LingBot-World (fast v2)~\citep{gao2026lingbot}
        & 79.4
        & 81.8
        & 76.8
        & 82.8
        & 86.5
        & 69.1 \\

        Kling 3.0~\citep{ying2026wbench}
        & 79.0
        & 81.4
        & 91.0
        & 69.4
        & 83.7
        & 69.3 \\

        LingBot-World (base-camera)~\citep{gao2026lingbot}
        & 78.5
        & 78.9
        & 72.6
        & 80.1
        & \textbf{89.9}
        & 71.2 \\

        Wan 2.7~\citep{wan2025wan}
        & 78.1
        & 81.5
        & \underline{91.4}
        & 64.4
        & 81.6
        & 71.8 \\

        HY-World 1.5 (AR-distill)~\citep{sun2025worldplay}
        & 78.1
        & 78.1
        & 72.2
        & 86.8
        & 86.9
        & 66.3 \\

        HY-Video 1.5~\citep{wu2025hunyuanvideo15}
        & 77.9
        & 77.6
        & 85.6
        & 71.4
        & 87.4
        & 67.4 \\

        LingBot-World (fast)~\citep{gao2026lingbot}
        & 77.4
        & 79.4
        & 77.9
        & 79.2
        & 84.9
        & 65.7 \\

        HappyOyster~\citep{ying2026wbench}
        & 76.8
        & 77.3
        & 74.2
        & 84.9
        & 84.3
        & 63.5 \\

        Lyra 2.0 (4-step AR)~\citep{shen2026lyra2}
        & 76.4
        & 77.1
        & 73.2
        & 85.6
        & 79.3
        & 66.7 \\

        Seedance 1.5~\citep{gao2025seedance}
        & 76.2
        & \underline{82.1}
        & 82.9
        & 66.3
        & 81.3
        & 68.4 \\

        Cosmos3-Super~\citep{nvidia2026cosmos3}
        & 76.0
        & 76.4
        & \textbf{91.6}
        & 58.0
        & 85.0
        & 69.2 \\

        SANA-WM (4-step AR)~\citep{zhu2026sanawm}
        & 76.0
        & 79.3
        & 76.1
        & 82.2
        & 80.7
        & 61.9 \\

        DreamX-World (5B AR)~\citep{team2026dreamx}
        & 75.0
        & 77.5
        & 80.8
        & 78.6
        & 74.9
        & 63.3 \\

        ABot-World~\citep{jiang2026abotworld}
        & 74.7
        & 76.8
        & 71.4
        & 84.0
        & 79.5
        & 61.7 \\

        Cosmos 2.5~\citep{cosmos_v1}
        & 74.6
        & 72.9
        & 83.3
        & 63.1
        & 86.5
        & 67.4 \\

        Cosmos3-Nano~\citep{nvidia2026cosmos3}
        & 74.2
        & 77.4
        & 87.3
        & 59.1
        & 83.7
        & 63.6 \\

        LTX-2.3~\citep{hacohen2026ltx2}
        & 74.2
        & 77.1
        & 85.2
        & 66.4
        & 77.2
        & 64.9 \\

        Genie 3~\citep{deepmind2025genie3}
        & 73.9
        & 75.2
        & 72.5
        & 73.4
        & 82.6
        & 65.7 \\

        \bottomrule
    \end{tabular}
    }
\end{table}


\begin{figure}[ht]
    \centering
    \adjustbox{
        trim=0 {.33\height} 0 0,
        clip,
        width=0.9\textwidth
      }{
        \includegraphics{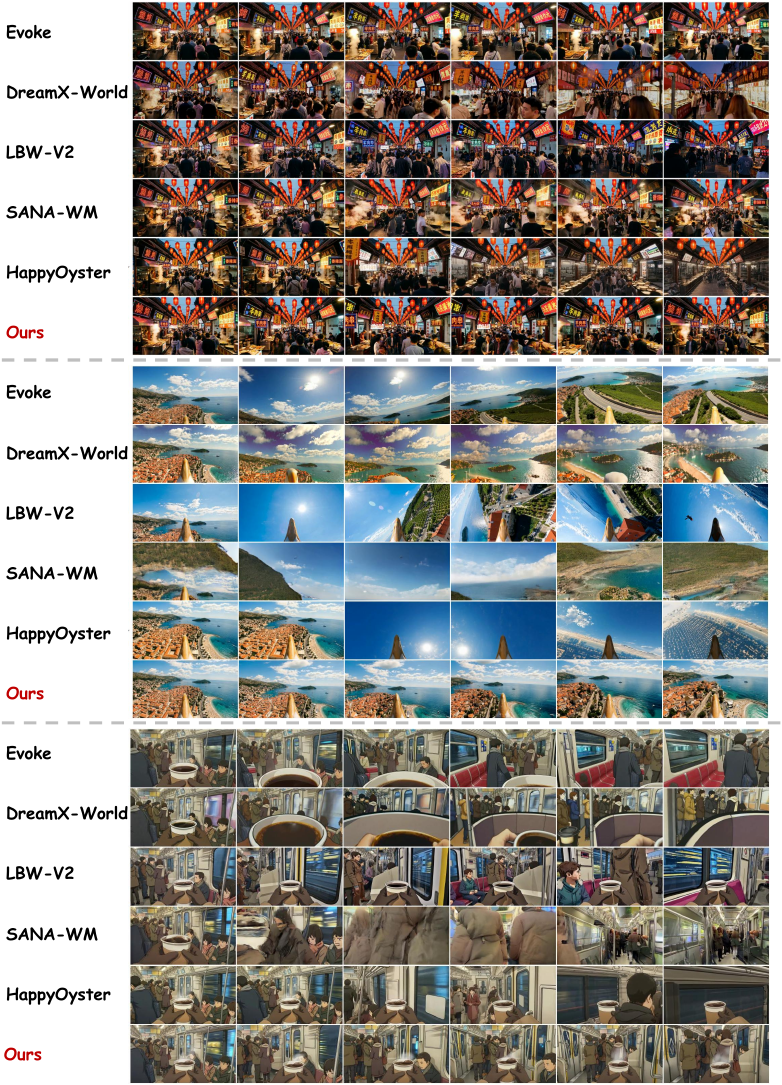}
      }
    \caption{WBench Navigation comparison.}
    \label{fig:wbench-comparison-2}
\end{figure}

\begin{figure}[ht]
    \centering
    \includegraphics[width=\textwidth,height=0.9\textheight,keepaspectratio]{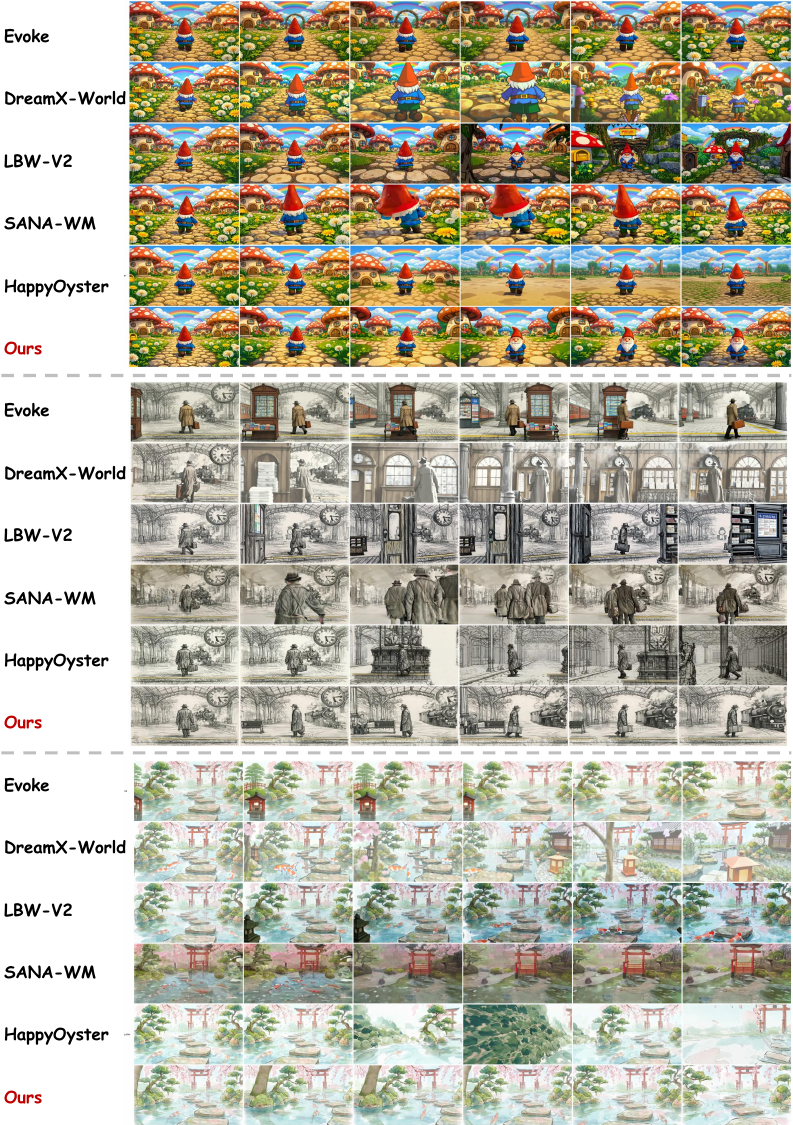}
    \caption{WBench Navigation comparison.}
    \label{fig:wbench-comparison-3}
\end{figure}

\begin{figure}[ht]
    \centering
    \includegraphics[width=\textwidth,height=0.9\textheight,keepaspectratio]{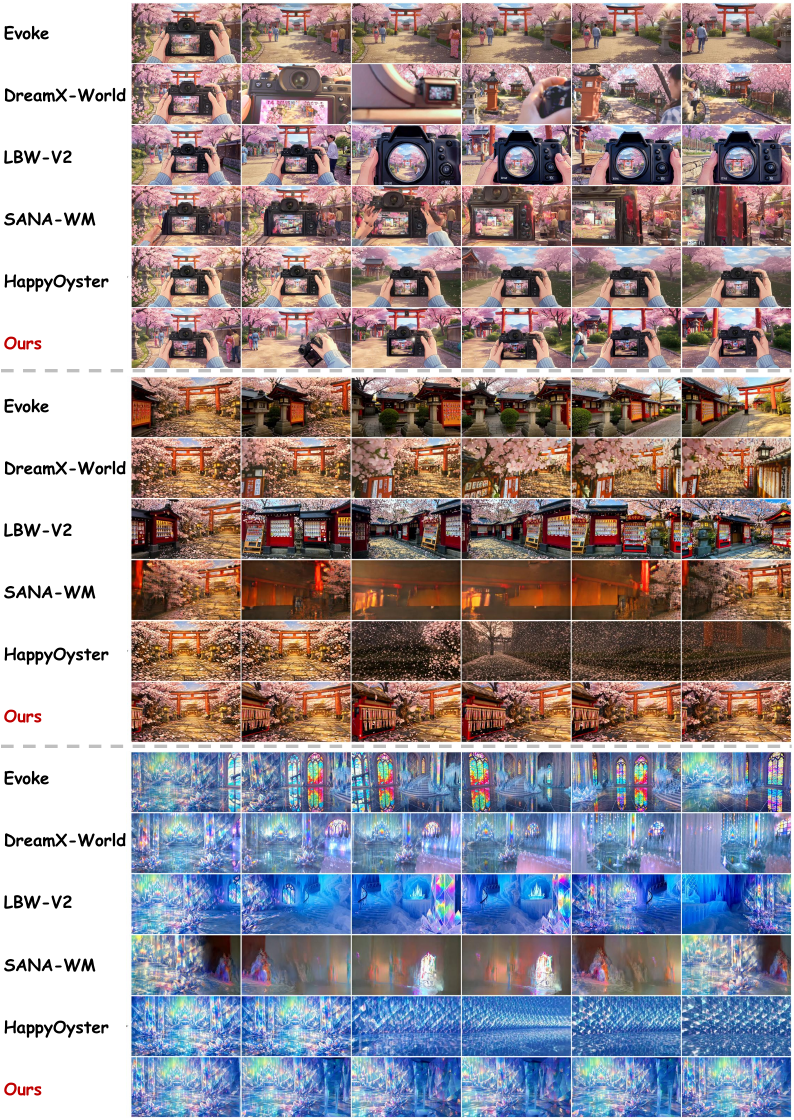}
    \caption{WBench Navigation comparison.}
    \label{fig:wbench-comparison-4}
\end{figure}


\begin{figure}[t!]
    \centering
    \includegraphics[
        width=\textwidth,
        height=0.92\textheight,
        keepaspectratio
    ]{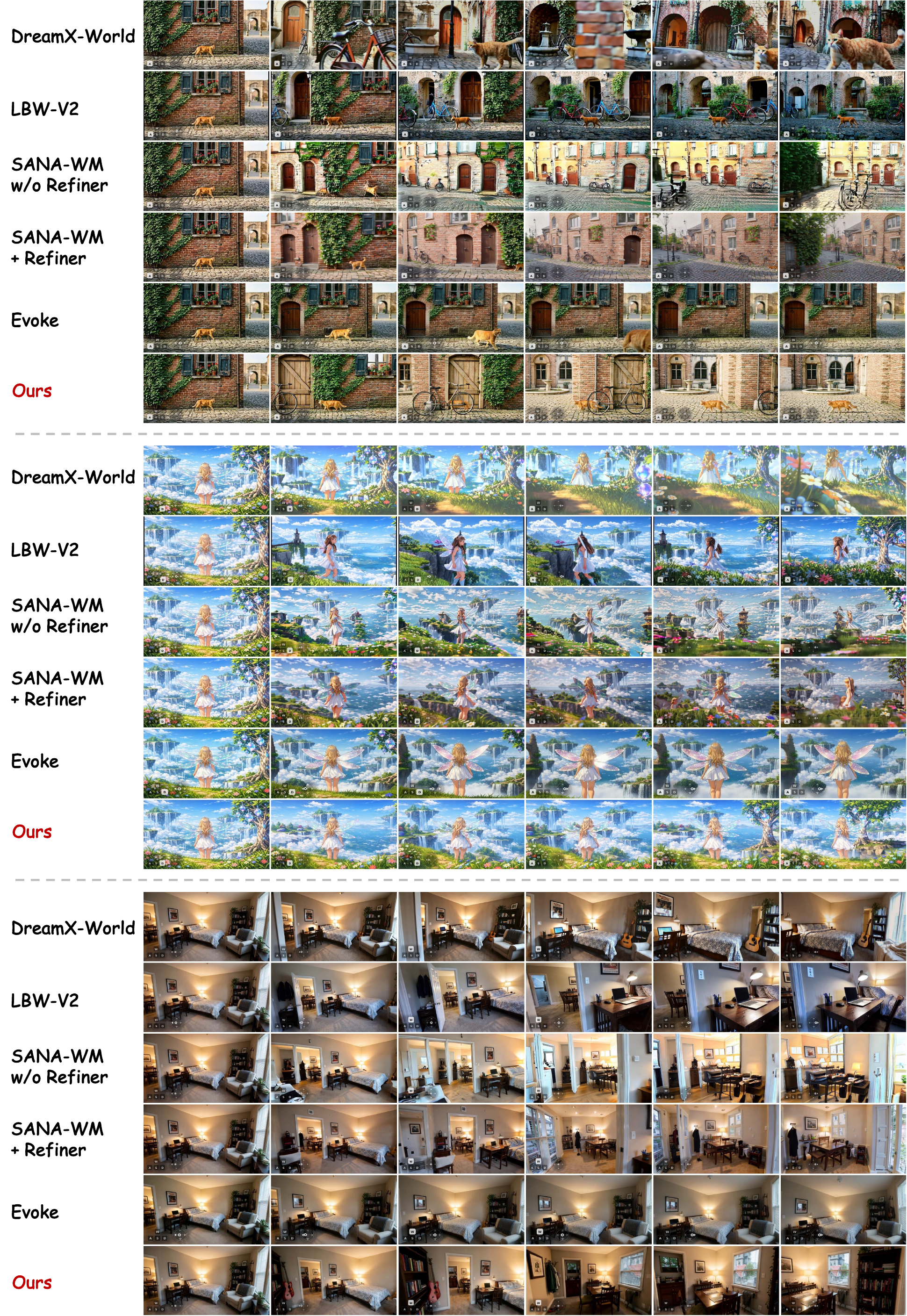}
    \caption{\textbf{Qualitative comparison of causal world models on WBench Navigation.}
    \model{}-Flash shows better action adherence and multi-turn consistency in
    scene structure, subject identity, and visual style.}
    \label{fig:wbench-causal-comparison}
\end{figure}

\noindent\textbf{Evaluation protocol.} We evaluate both \model and its distilled four-step causal variant, \model{}-Flash, on the official 158-case Navigation split of WBench~\citep{ying2026wbench}, covering first- and third-person interactive navigation. We convert the supported control interfaces to our shared internal trajectory condition and follow the official multi-turn evaluation protocol. We report all five official dimensions and their overall Average: \emph{Quality} measures perceptual video quality; \emph{Setting} measures scene and subject adherence; \emph{Interaction} measures navigation and action following; \emph{Consistency} measures spatial, temporal, and viewpoint persistence; and \emph{Physical} measures causal fidelity and visual plausibility. All scores follow the official 0--100 normalization and aggregation, with higher values indicating better performance.

\noindent\textbf{Quantitative results.}
As shown in \cref{tab:wbench-navigation}, both variants rank at the top of the
WBench Navigation benchmark, demonstrating strong interactive world-modeling
capability before and after causal distillation.
The undistilled multi-step \model achieves the highest overall Average score of
81.7 and a strong Consistency score of 89.8, showing navigation
performance together with stable world-state preservation across multi-turn
interactions.
Building on this strong base model, \model{}-Flash closely follows with an
Average score of 81.0 and achieves the highest Interaction score of
87.9, slightly surpassing the undistilled model at 87.2.
Despite replacing bidirectional multi-step generation with causal four-step
inference, \model{}-Flash therefore retains most of the original model's overall
capability while preserving, and even slightly improving, control responsiveness.
Together, these results establish \model as a strong interactive world model and
show that the proposed distillation effectively transfers its capability to the
more efficient causal few-step variant.

\noindent\textbf{Qualitative results.}
As illustrated in \cref{fig:wbench-comparison-2,fig:wbench-comparison-4}, competing world models often exhibit abrupt viewpoint changes, unintended zooming, subject-scale variation, or scene-layout drift under the same initial observation and navigation condition. In contrast, \model{} more faithfully follows the prescribed navigation while
better preserving scene structure, subject appearance, and visual style across
successive interactions. As shown in
\cref{fig:wbench-causal-comparison}, we further evaluate \model{}-Flash under
repeated interactions and extended autoregressive rollout using three
representative multi-turn cases. The action sequences are
\texttt{A} $\rightarrow$ \texttt{A} $\rightarrow$
\texttt{A} $\rightarrow$ \texttt{A},
\texttt{Left} $\rightarrow$ \texttt{Left} $\rightarrow$
\texttt{Right} $\rightarrow$ \texttt{Right}, and
\texttt{Left} $\rightarrow$ \texttt{W} $\rightarrow$
\texttt{Right}, respectively. Here, \texttt{Left}, \texttt{Right},
\texttt{Up}, and \texttt{Down} control camera rotation, while
\texttt{W}, \texttt{S}, \texttt{A}, and \texttt{D} control camera
translation. Compared with representative recent causal world models,
\model{}-Flash follows these turn-wise control changes more consistently while
exhibiting less accumulated drift in viewpoint, scene layout, subject identity,
and visual style. Competitive baselines such as LingBot2-Fast and Evoke remain
coherent over multiple turns, but still show noticeable action deviation and
gradual degradation in subject identity or visual style as the rollout
progresses. Overall, \model{}-Flash better preserves both controllability and
visual consistency over extended autoregressive generation, indicating
improved robustness to long-horizon error accumulation.


\FloatBarrier

\subsection{SANA-WM-Bench}
\label{sec:sana-wm-bench}

\begin{table}[ht]
    \centering
    \small
    \setlength{\tabcolsep}{9pt}
    \renewcommand{\arraystretch}{1.06}
    \caption{Short-horizon comparison on SANA-WM-Bench using the front 241 frames of each trajectory.
    Each split contains 80 scenes.
    VBench Overall follows the benchmark convention of reporting normalized visual quality on a 0--100 scale.
    Camera accuracy reports rotation error $R$ in degrees, relative translation error $T$, and camera-motion consistency error CMC after similarity alignment.}
    \label{tab:sana-wm-bench-241}
    \begin{tabular}{lcccc}
        \toprule
        Method & VBench Overall $\uparrow$ & $R\downarrow$ & $T\downarrow$ & CMC$\downarrow$ \\
        \midrule
        \multicolumn{5}{l}{\textit{Simple-Trajectory Split}} \\
        Matrix-Game 3~\citep{wang2026matrixgame3}       & 79.41 & 4.474 & 0.406 & 0.453 \\
        LingBot-World-v2~\citep{gao2026lingbot}         & 81.13 & 3.092 & 0.282 & 0.310 \\
        HY-WorldPlay~\citep{sun2025worldplay}            & 82.05 & 2.256 & 0.284 & 0.298 \\
        DreamX-World~\citep{team2026dreamx}              & \underline{82.42} & 1.839 & 0.261 & 0.273 \\
        SANA-WM~\citep{zhu2026sanawm}                    & 81.75 & \textbf{0.489} & \underline{0.194} & \underline{0.195} \\
        \textbf{\model}    & \textbf{83.91} & \underline{0.523} & \textbf{0.160} & \textbf{0.162} \\
        \midrule
        \multicolumn{5}{l}{\textit{Hard-Trajectory Split}} \\
        Matrix-Game 3~\citep{wang2026matrixgame3}       & 79.37 & 12.819 & 0.583 & 0.717 \\
        LingBot-World-v2~\citep{gao2026lingbot}         & 81.67 & 6.367 & 0.357 & 0.426 \\
        HY-WorldPlay~\citep{sun2025worldplay}            & 81.24 & 6.346 & 0.387 & 0.437 \\
        DreamX-World~\citep{team2026dreamx}              & \underline{82.03} & 8.804 & 0.467 & 0.556 \\
        SANA-WM~\citep{zhu2026sanawm}                    & 81.79 & \textbf{1.248} & \textbf{0.199} & \textbf{0.206} \\
        \textbf{\model}    & \textbf{83.96} & \underline{1.697} & \underline{0.256} & \underline{0.266} \\
        \bottomrule
    \end{tabular}
\end{table}

\begin{table}[ht]
    \centering
    \scriptsize
    \setlength{\tabcolsep}{3.2pt}
    \renewcommand{\arraystretch}{1.06}
    \caption{Long-horizon comparison on the full SANA-WM-Bench trajectory.
    Revisit consistency is measured on same-pose pairs.
    VBench Overall is the normalized Quality component.
    Temporal degradation is $\Delta\mathrm{IQ}=\mathrm{IQ}_{0\text{--}10}-\mathrm{IQ}_{50\text{--}60}$ in percentage points; lower is better, and a negative value indicates improvement in the last window.
    Best and second-best results within each split are shown in bold and underlined, respectively.}
    \label{tab:sana-wm-bench-961}
    \begin{tabular}{lcccccccc}
        \toprule
        & \multicolumn{4}{c}{Trajectory and visual quality} & \multicolumn{3}{c}{Revisit consistency} & Temporal \\
        \cmidrule(lr){2-5}\cmidrule(lr){6-8}\cmidrule(lr){9-9}
        Method & VBench$\uparrow$ & $R\downarrow$ & $T\downarrow$ & CMC$\downarrow$ & PSNR$\uparrow$ & SSIM$\uparrow$ & LPIPS$\downarrow$ & $\Delta$IQ$\downarrow$ \\
        \midrule
        \multicolumn{9}{l}{\textit{Simple-Trajectory Split}} \\
        Infinite-World~\citep{wu2026infiniteworld}       & 79.18 & 16.55 & 1.98  & 2.08  & 12.60 & 0.284 & 0.595 &  6.72 \\
        LingBot-World~\citep{gao2026lingbot}             & \textbf{81.82} & 10.47 & 2.01  & 2.05  & \underline{14.59} & \textbf{0.366} & \textbf{0.394} & \underline{0.04} \\
        HY-WorldPlay~\citep{sun2025worldplay}            & 68.82 & 17.89 & 2.36  & 2.45  & 12.83 & 0.321 & 0.616 & 23.59 \\
        Matrix-Game 3~\citep{wang2026matrixgame3}       & 78.53 & 12.96 & 1.83  & 1.92  & 12.29 & 0.326 & 0.553 &  2.41 \\
        SANA-WM~\citep{zhu2026sanawm}                    & 79.29 & \underline{7.59} & \underline{1.59} & \underline{1.63} & 14.16 & 0.333 & 0.458 &  3.79 \\
        \textbf{\model} & \underline{81.36} & \textbf{3.22} & \textbf{0.918} & \textbf{0.927} & \textbf{15.10} & \underline{0.335} & \underline{0.451} & \textbf{0.02} \\
        \midrule
        \multicolumn{9}{l}{\textit{Hard-Trajectory Split}} \\
        Infinite-World~\citep{wu2026infiniteworld}       & 79.51 & 41.31 & 2.49  & 2.84  & 12.04 & 0.248 & 0.617 &  4.16 \\
        LingBot-World~\citep{gao2026lingbot}             & \textbf{81.89} & 18.99 & \underline{1.65} & 1.81 & 14.08 & \underline{0.332} & \textbf{0.436} & \underline{0.58} \\
        HY-WorldPlay~\citep{sun2025worldplay}            & 70.46 & 35.46 & 2.34  & 2.64  & 13.72 & 0.328 & 0.654 & 25.88 \\
        Matrix-Game 3~\citep{wang2026matrixgame3}       & 78.79 & 18.79 & 1.67  & 1.82  & 12.17 & 0.317 & 0.556 & \textbf{0.32} \\
        SANA-WM~\citep{zhu2026sanawm}                    & 79.60 & \textbf{10.02} & 1.66  & \underline{1.72} & \underline{14.10} & 0.327 & \underline{0.469} &  3.09 \\
        \textbf{\model} & \underline{81.72} & \underline{12.05} & \textbf{1.265} & \textbf{1.358} & \textbf{14.58} & \textbf{0.333} & 0.482 & 0.93 \\
        \bottomrule
    \end{tabular}
\end{table}

\begin{table}[ht]
    \centering
    \scriptsize
    \setlength{\tabcolsep}{3.2pt}
    \renewcommand{\arraystretch}{1.06}
    \caption{
    Long-horizon comparison of causal world models on SANA-WM-Bench under the
    961-frame protocol.
    \model{}-Flash denotes our four-step causal streaming model.
    Revisit consistency is measured on same-pose pairs, and
    $\Delta\mathrm{IQ}$ measures temporal imaging-quality degradation.
    Best and second-best results within each split are shown in bold and
    underlined, respectively.
    }
    \label{tab:sana-wm-bench-flash-961}
    \begin{tabular}{lcccccccc}
        \toprule
        & \multicolumn{4}{c}{Trajectory and visual quality}
        & \multicolumn{3}{c}{Revisit consistency}
        & Temporal \\
        \cmidrule(lr){2-5}
        \cmidrule(lr){6-8}
        \cmidrule(lr){9-9}
        Method
        & VBench$\uparrow$
        & $R\downarrow$
        & $T\downarrow$
        & CMC$\downarrow$
        & PSNR$\uparrow$
        & SSIM$\uparrow$
        & LPIPS$\downarrow$
        & $\Delta$IQ$\downarrow$ \\
        \midrule

        \multicolumn{9}{l}{\textit{Simple-Trajectory Split}} \\

        Evoke~\citep{yin2026alaya}
        & \underline{80.11}
        & \underline{9.859}
        & \underline{2.159}
        & \underline{2.190}
        & \underline{13.50}
        & \underline{0.2774}
        & \underline{0.5010}
        & \underline{-0.49} \\

        LingBot-World-v2~\citep{gao2026lingbot}
        & 79.20
        & 22.516
        & 2.391
        & 2.541
        & 11.90
        & 0.1715
        & 0.5906
        & 2.59 \\

        SANA-WM + Causal Refiner~\citep{zhu2026sanawm}
        & 79.55
        & 17.875
        & 2.222
        & 2.338
        & \textbf{14.41}
        & 0.2660
        & 0.5748
        & \textbf{-0.75} \\

        SANA-WM w/o Refiner~\citep{zhu2026sanawm}
        & 78.21
        & 15.359
        & 2.288
        & 2.393
        & 9.19
        & 0.1661
        & 0.6321
        & -0.12 \\

        \midrule

        \textcolor{EchoAccent}{\textbf{\model{}-Flash}}
        & \textbf{80.13}
        & \textbf{5.009}
        & \textbf{1.592}
        & \textbf{1.611}
        & \textbf{14.41}
        & \textbf{0.2998}
        & \textbf{0.4802}
        & 2.09 \\

        \midrule
        \multicolumn{9}{l}{\textit{Hard-Trajectory Split}} \\

        Evoke~\citep{yin2026alaya}
        & \underline{80.75}
        & 20.097
        & \underline{1.924}
        & \underline{2.060}
        & 13.06
        & 0.2592
        & \underline{0.5292}
        & \underline{0.30} \\

        LingBot-World-v2~\citep{gao2026lingbot}
        & 78.74
        & 34.416
        & 2.125
        & 2.413
        & 12.34
        & 0.2183
        & 0.5732
        & 3.84 \\

        SANA-WM + Causal Refiner~\citep{zhu2026sanawm}
        & 79.60
        & 23.794
        & 1.932
        & 2.124
        & \underline{13.89}
        & \underline{0.2605}
        & 0.5696
        & \textbf{-0.24} \\

        SANA-WM w/o Refiner~\citep{zhu2026sanawm}
        & 78.27
        & \underline{18.717}
        & 2.006
        & 2.142
        & 9.26
        & 0.1729
        & 0.6077
        & 2.06 \\

        \midrule

        \textcolor{EchoAccent}{\textbf{\model{}-Flash}}
        & \textbf{81.06}
        & \textbf{14.221}
        & \textbf{1.735}
        & \textbf{1.829}
        & \textbf{14.04}
        & \textbf{0.2910}
        & \textbf{0.4893}
        & 1.30 \\

        \bottomrule
    \end{tabular}
\end{table}

\begin{figure}[t!]
    \centering
    \includegraphics[width=\textwidth]{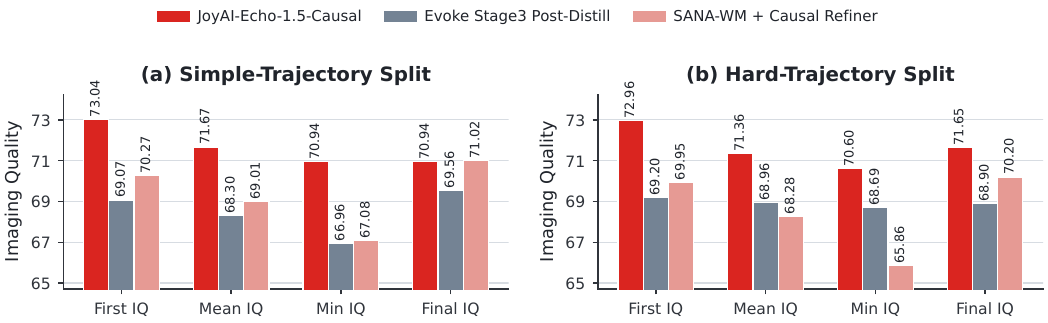}
    \caption{Imaging quality comparison on the Simple- and Hard-Trajectory splits. We report the first-window, mean, minimum, and final-window IQ. Our method achieves consistently higher mean and minimum IQ than Evoke and SANA-WM with Causal Refiner. Although the competing methods exhibit smaller IQ drops, their overall IQ remains lower, indicating that a small drop can also result from consistently low imaging quality rather than stronger long-horizon quality preservation.}
    \label{fig:iq-comparison}
\end{figure}

\begin{figure}[!t]
    \centering
    \includegraphics[
  width=0.96\textwidth,
  height=0.80\textheight,
  keepaspectratio
]{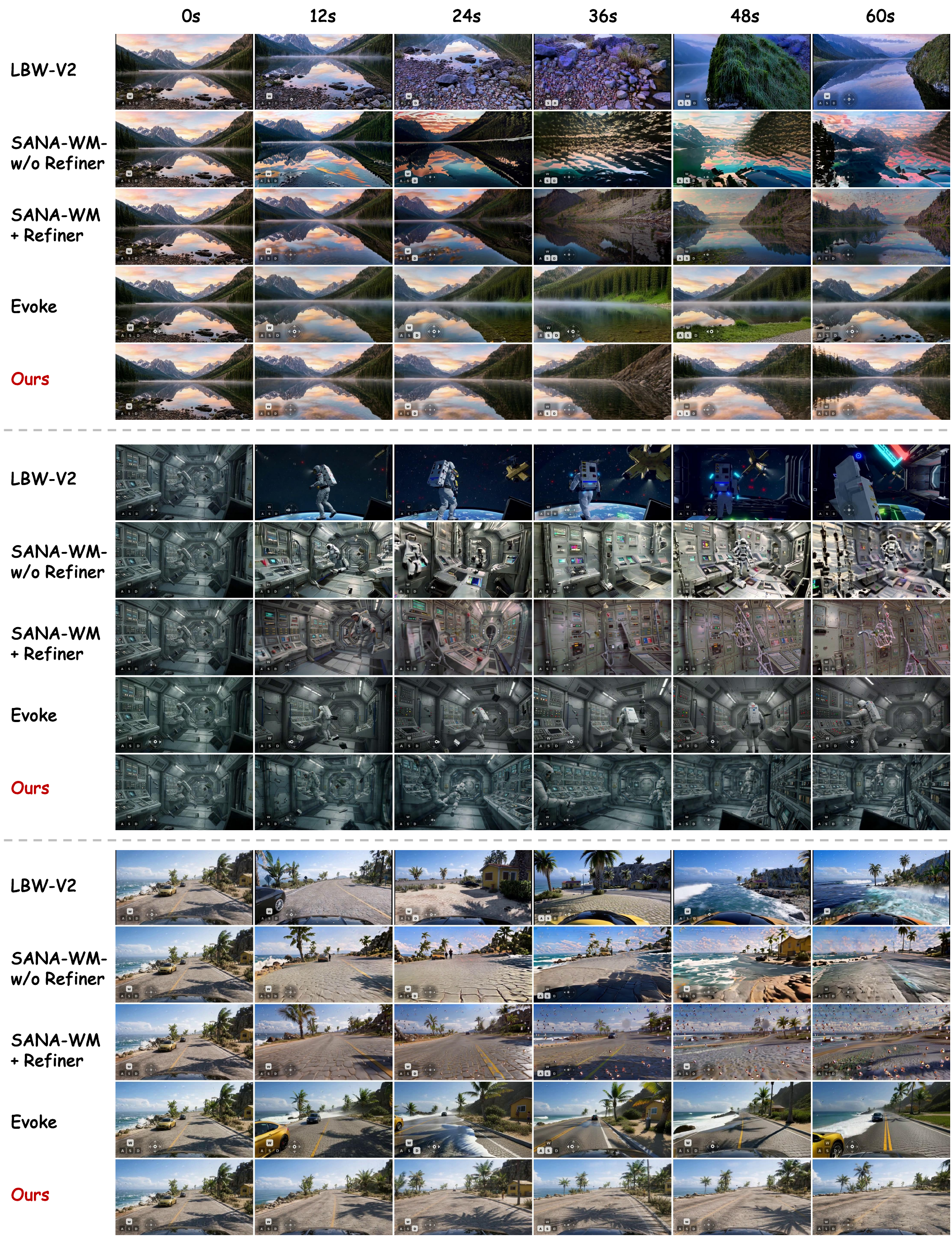}
    \caption{
    Qualitative comparison of long-horizon causal streaming on representative
    SANA-WM-Bench trajectories under the 961-frame protocol.
    We compare \model{}-Flash with LingBot-World-v2 (LBW-V2), SANA-WM with and
    without Causal Refiner, and Evoke, using the same initial observation and
    camera-trajectory condition.
    Frames are shown at 12-second intervals.
    \model{}-Flash exhibits less accumulated autoregressive drift while better
    preserving trajectory adherence, scene identity, visual style, and major
    spatial structure over the extended rollout.
    }
    \label{fig:sana-causal-qualitative}
\end{figure}

\noindent\textbf{Evaluation protocol.}
We evaluate SANA-WM-Bench~\citep{zhu2026sanawm} under three complementary
settings:
(1) a 241-frame short-horizon evaluation of the undistilled multi-step model,
(2) the official 961-frame long-horizon
protocol, where four overlapping 241-frame generations are sequentially chained,
and
(3) \model{}-Flash under the same 961-frame horizon using four-step causal
streaming inference.
The first setting provides a short-horizon reference for trajectory following
and visual quality, while the latter two characterize long-horizon generation
before and after causal few-step distillation under their respective inference
schemes.

Following the benchmark protocol, we report rotation error ($R$, in degrees),
relative translation error ($T$), camera-motion consistency error (CMC), and
VBench Overall.
For the 961-frame evaluations, we additionally measure viewpoint-revisit
consistency using PSNR, SSIM, and LPIPS between frames that revisit
approximately the same camera pose, and report temporal imaging-quality
degradation,
$\Delta\mathrm{IQ}=q_1-q_6$,
where $q_1$ and $q_6$ denote the VBench Imaging Quality scores of the first
and final 10-second windows, respectively.


\noindent\textbf{Quantitative results.}
As shown in \cref{tab:sana-wm-bench-241}, \model achieves the highest
241-frame VBench Overall on both Simple and Hard splits, with 83.91 and 83.96,
respectively.
On Simple trajectories, it also achieves the lowest translation and CMC errors
while remaining close to SANA-WM in rotation accuracy; on Hard trajectories,
SANA-WM obtains lower pose errors, whereas \model maintains the strongest
visual quality.

Under the 961-frame long-horizon protocol
(\cref{tab:sana-wm-bench-961}), the undistilled multi-step \model remains
competitive in trajectory following while preserving strong visual quality and
scene consistency.
On the Simple split, it achieves a VBench Overall of 81.36, the lowest
translation and CMC errors, and the highest revisit PSNR of 15.10\,dB.
Its rotation error increases from $0.523^\circ$ at 241 frames to
$3.223^\circ$ at 961 frames, and further reaches $12.05^\circ$ on the Hard
split, indicating that accumulated pose drift remains the main challenge under
difficult long-horizon trajectories.

We further compare \model{}-Flash with recent causal world models under the
same 961-frame horizon.
As shown in \cref{tab:sana-wm-bench-flash-961}, \model{}-Flash achieves the
highest VBench Overall on both Simple and Hard splits, with 80.13 and 81.06,
respectively, together with the strongest trajectory accuracy and revisit
consistency among the evaluated causal methods.
To further examine quality degradation over long rollouts,
\cref{fig:iq-comparison} summarizes the Imaging Quality (IQ) across temporal
windows.
\model{}-Flash maintains substantially higher initial, mean, and minimum IQ
than competing methods while remaining competitive in the final window.
Although some baselines exhibit a smaller $\Delta\mathrm{IQ}$, their lower
absolute IQ suggests that temporal degradation should be interpreted jointly
with the overall quality level.
Overall, four-step causal distillation introduces only moderate degradation
relative to the undistilled model while largely preserving long-horizon visual
quality and control fidelity.

\noindent\textbf{Qualitative results.}
The long-horizon causal streaming comparisons in
\cref{fig:sana-causal-qualitative} further highlight differences in
autoregressive error accumulation.
Competing causal models progressively exhibit viewpoint deviation, scene-layout
drift, or changes in appearance and visual style as the streaming rollout proceeds.
In contrast, \model{}-Flash better preserves scene and subject identity, visual
style, and major geometric structures while following the prescribed camera
trajectory.
This sustained consistency at later timestamps is aligned with the quantitative
trajectory, revisit-consistency, and temporal-quality results, demonstrating
stronger robustness to long-horizon autoregressive drift.



\subsection{User Study}

\begin{table}[ht]
    \centering
    \small
    \setlength{\tabcolsep}{4.2pt}
    \caption{Aggregate pairwise user-study judgments over 200 cases. Each cell reports the percentage of judgments. ``Both good'' and ``both bad'' preserve the two distinct same-quality outcomes.}
    \label{tab:user-study-overall}
    \resizebox{\textwidth}{!}{%
    \begin{tabular}{@{}llcccc@{}}
        \toprule
        Competitor & Dimension & Competitor preferred & Both good & Both bad & \model preferred \\
        \midrule
        \multirow{6}{*}{LingBot-World-v2}
        & Semantic following             & 15.14\% & 24.93\% & 25.93\% & \textbf{34.00\%} \\
        & Initial-world consistency      & 7.71\%  & 62.36\% & 15.29\% & \textbf{14.64\%} \\
        & Motion                         & 20.00\% & 13.71\% & 41.00\% & \textbf{25.29\%} \\
        & Spatial-temporal consistency   & 14.71\% & 27.36\% & 25.29\% & \textbf{32.64\%} \\
        & Visual aesthetics              & 19.93\% & 28.00\% & 19.57\% & \textbf{32.50\%} \\
        & Overall preference             & 21.57\% & 4.21\%  & 27.71\% & \textbf{46.50\%} \\
        \midrule
        \multirow{6}{*}{HappyOyster}
        & Semantic following             & 29.44\% & 51.50\% & 5.44\%  & 13.63\% \\
        & Initial-world consistency      & 15.44\% & 42.81\% & 2.25\%  & \textbf{39.50\%} \\
        & Motion                         & 24.56\% & 11.88\% & 44.63\% & 18.94\% \\
        & Spatial-temporal consistency   & 10.44\% & 18.19\% & 13.69\% & \textbf{57.69\%} \\
        & Visual aesthetics              & 23.81\% & 1.13\%  & 22.19\% & \textbf{52.88\%} \\
        & Overall preference             & 27.06\% & 2.19\%  & 7.63\%  & \textbf{63.13\%} \\
        \bottomrule
    \end{tabular}%
    }
\end{table}

\begin{table}[t]
    \centering
    \footnotesize
    \setlength{\tabcolsep}{3.8pt}
    \caption{
    Scene-level user-study results against LingBot-World-v2 and HappyOyster.
    All values are percentages. The LingBot-World-v2 comparison uses seven annotators,
    while the HappyOyster comparison uses eight annotators.
    }
    \label{tab:user-study-scenes}
    \resizebox{0.98\textwidth}{!}{%
    \begin{tabular}{llcccccccc}
        \toprule
        & &
        \multicolumn{4}{c}{\textbf{vs. LingBot-World-v2}} &
        \multicolumn{4}{c}{\textbf{vs. HappyOyster}} \\
        \cmidrule(lr){3-6}
        \cmidrule(lr){7-10}

        Scene & Dimension
        & LBW-v2 & Both good & Both bad & \model
        & HappyOyster & Both good & Both bad & \model \\
        \midrule

        \multirow{6}{*}{Autonomous driving}
        & Semantic following
        & 30.48 & 29.52 & 17.14 & 22.86
        & 18.33 & 44.17 & 13.33 & 24.17 \\

        & Initial-world consistency
        & 3.81 & 66.67 & 15.24 & 14.29
        & 10.83 & 50.00 & 0.00 & 39.17 \\

        & Motion
        & 21.90 & 6.67 & 42.86 & 28.57
        & 5.83 & 10.83 & 35.83 & 47.50 \\

        & Spatial-temporal consistency
        & 10.48 & 31.43 & 29.52 & 28.57
        & 10.00 & 24.17 & 12.50 & 53.33 \\

        & Visual aesthetics
        & 11.43 & 34.29 & 21.90 & 32.38
        & 19.17 & 3.33 & 17.50 & 60.00 \\

        & Overall preference
        & 24.76 & 1.90 & 30.48 & \textbf{42.86}
        & 14.17 & 3.33 & 7.50 & \textbf{75.00} \\

        \midrule

        \multirow{6}{*}{Embodied robot}
        & Semantic following
        & 12.90 & 9.45 & 42.17 & 35.48
        & 42.94 & 41.53 & 6.85 & 8.67 \\

        & Initial-world consistency
        & 7.37 & 60.60 & 21.66 & 10.37
        & 21.37 & 42.14 & 2.02 & 34.48 \\

        & Motion
        & 21.20 & 7.83 & 53.46 & 17.51
        & 31.45 & 7.46 & 52.62 & 8.47 \\

        & Spatial-temporal consistency
        & 15.90 & 25.58 & 34.10 & 24.42
        & 10.89 & 17.74 & 18.15 & 53.23 \\

        & Visual aesthetics
        & 22.81 & 28.34 & 24.42 & 24.42
        & 30.44 & 1.01 & 29.64 & 38.91 \\

        & Overall preference
        & 22.35 & 3.00 & 39.63 & \textbf{35.02}
        & 40.93 & 1.81 & 11.69 & \textbf{45.56} \\

        \midrule

        \multirow{6}{*}{Multi-social}
        & Semantic following
        & 21.43 & 28.57 & 14.29 & 35.71
        & 12.50 & 62.50 & 6.25 & 18.75 \\

        & Initial-world consistency
        & 21.43 & 64.29 & 7.14 & 7.14
        & 6.25 & 62.50 & 0.00 & 31.25 \\

        & Motion
        & 14.29 & 42.86 & 14.29 & 28.57
        & 0.00 & 31.25 & 43.75 & 25.00 \\

        & Spatial-temporal consistency
        & 21.43 & 42.86 & 0.00 & 35.71
        & 0.00 & 12.50 & 6.25 & 81.25 \\

        & Visual aesthetics
        & 28.57 & 28.57 & 7.14 & 35.71
        & 18.75 & 0.00 & 12.50 & 68.75 \\

        & Overall preference
        & 28.57 & 7.14 & 7.14 & \textbf{57.14}
        & 12.50 & 0.00 & 6.25 & \textbf{81.25} \\

        \midrule

        \multirow{6}{*}{Natural outdoor}
        & Semantic following
        & 17.86 & 16.43 & 37.86 & 27.86
        & 46.88 & 43.75 & 6.88 & 2.50 \\

        & Initial-world consistency
        & 6.43 & 67.14 & 13.57 & 12.86
        & 21.88 & 34.38 & 1.25 & 42.50 \\

        & Motion
        & 15.71 & 7.86 & 51.43 & 25.00
        & 38.75 & 10.00 & 41.25 & 10.00 \\

        & Spatial-temporal consistency
        & 15.71 & 21.43 & 26.43 & 36.43
        & 13.13 & 14.37 & 8.13 & 64.38 \\

        & Visual aesthetics
        & 14.29 & 28.57 & 22.86 & 34.29
        & 28.75 & 0.63 & 21.88 & 48.75 \\

        & Overall preference
        & 17.86 & 4.29 & 31.43 & \textbf{46.43}
        & 41.25 & 1.25 & 8.75 & \textbf{48.75} \\

        \bottomrule
    \end{tabular}%
    }
\end{table}

We conduct subjective evaluation on a self-built World-Model Benchmark (WMB) containing 200 cases.
WMB is designed to measure overall world-model behavior under interactive generation.

\noindent{\textbf{Benchmark composition.}}
The cases cover five scene groups: 100 game scenes, 50 humanoid-robot scenes, 20 natural outdoor scenes, 15 household indoor scenes, and 15 driving scenes.
The viewpoint split contains 132 third-person cases and 68 first-person cases.
This composition stresses both game-oriented interaction and generalization to embodied, natural, household, and driving environments.

\noindent{\textbf{Control protocol.}}
WMB follows the WBench control convention~\citep{ying2026wbench}.
For first-person cases, WASD translates the camera and the arrow keys rotate the viewpoint.
For third-person cases, WASD moves the visible subject and the arrow keys orbit the camera around that subject.
The same command semantics are used for every model under comparison, while the generated video is evaluated from the resulting reference, text, and control condition.

\noindent{\textbf{Subjective dimensions.}}
Annotators score five dimensions: (1) semantic following, measuring adherence to the requested scene and behavior; (2) initial-world consistency, measuring preservation of the reference observation and initial world state; (3) spatial-temporal consistency, measuring geometry, identity, and layout stability over time; (4) motion quality, measuring smoothness, responsiveness, and physical plausibility; and (5) visual aesthetics, measuring composition, lighting, texture, and overall perceptual quality.

Table~\ref{tab:user-study-overall} shows that \model receives the largest share of overall-preference votes against both competitors: 46.50\% against LingBot-World-v2 and 63.13\% against HappyOyster.
Against LingBot-World-v2, \model also leads semantic following (34.00\%), spatial-temporal consistency (32.64\%), and visual aesthetics (32.50\%), whereas initial-world consistency is dominated by both-good judgments (62.36\%) and motion most often receives both-bad judgments (41.00\%).
Against HappyOyster, the clearest advantages are spatial-temporal consistency (57.69\%) and visual aesthetics (52.88\%).
Semantic following remains a weakness in this comparison: HappyOyster is preferred in 29.44\% of judgments versus 13.63\% for \model, with 51.50\% of pairs judged both good.

The reported scene subset contains 99 cases across autonomous driving, embodied robot, multi-social, and natural outdoor settings.
For overall preference, \model leads LingBot-World-v2 in all four groups, with vote shares ranging from 35.02\% in embodied robot to 57.14\% in multi-social scenes.
Against HappyOyster, \model is strongly preferred in autonomous driving (75.00\%) and multi-social scenes (81.25\%), while the margins are narrower in embodied robot (45.56\% versus 40.93\%) and natural outdoor scenes (48.75\% versus 41.25\%).
The dimension-level breakdown also exposes localized failure modes: against HappyOyster, \model receives only 8.67\% of semantic-following votes in embodied-robot scenes and 2.50\% in natural-outdoor scenes, despite leading those scenes in overall preference.


\begin{figure}[!t]
  \centering
  \adjustbox{
    trim=0 {.4\height} 0 0,
    clip,
    width=0.9\textwidth
  }{
    \includegraphics{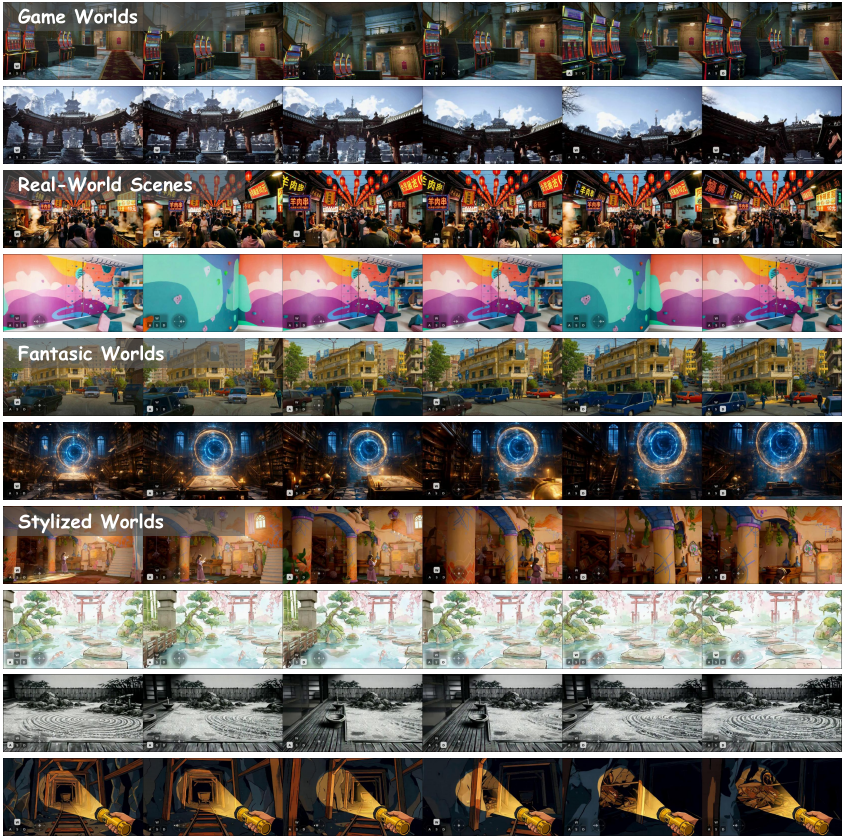}
  }
  \caption{\textbf{First-person world generalization.}
  Temporally ordered rollouts cover game worlds, real-world scenes,
  fantastic worlds, and stylized worlds.
  Across varied layouts and rendering styles, \model uses the same
  camera-intent interface for observer-centric navigation.}
  \label{fig:qualitative-fpp}
\end{figure}
\begin{figure}[t!]
    \centering
    \includegraphics[width=0.9\textwidth]{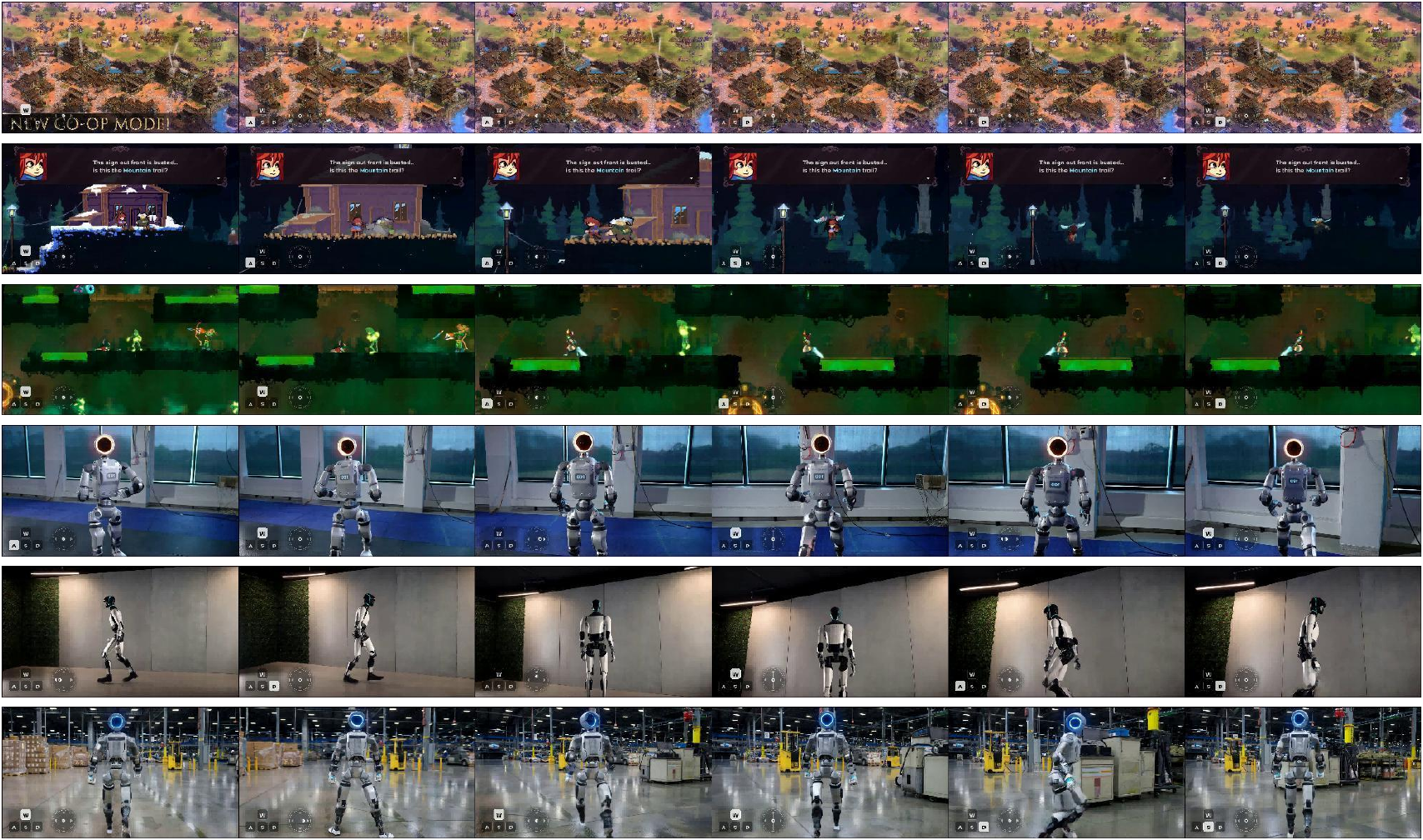}
    \caption{\textbf{Generalization beyond the training domains.}
    Our training mixture contains no 2D or 2.5D game data and no robot data.
    The top three rows show 2D/2.5D game scenes and the bottom three rows show
    robot environments; columns are temporally ordered frames under navigation
    controls.
    \model preserves coherent viewpoint evolution and scene or subject
    structure across these unseen domains.}
    \label{fig:generalization}
\end{figure}

\subsection{Qualitative Analysis}
\label{sec:qualitative-analysis}

Quantitative benchmarks measure performance under predefined protocols, but do not fully reveal the breadth of an enterable omnimodal world.
We therefore organize the qualitative analysis along five complementary capability axes: first-person generalization across worlds, third-person interaction across visible subjects, spatial structure recoverable from generated views, native synchronized generation of video and audio, and multi-turn preservation of audio-visual context.
Together, these axes examine whether the same camera-intent interface remains effective as the world, viewpoint, subject, spatial evidence, output modality, and interaction horizon change.

\noindent{\textbf{First-Person World Generalization.}}
We first examine whether one camera-intent interface generalizes across substantially different world distributions.
Figure~\ref{fig:qualitative-fpp} covers game-rendered worlds, real-world scenes, fantastic environments, and stylized content under the same first-person trajectory representation.
Each row shows temporally ordered output frames, allowing changes in world appearance and layout to be examined without changing the interaction interface.
The examples demonstrate observer-centric navigation across diverse world distributions without a domain-specific controller.

\begin{figure}[t!]
    \centering
    \includegraphics[width=0.9\textwidth]{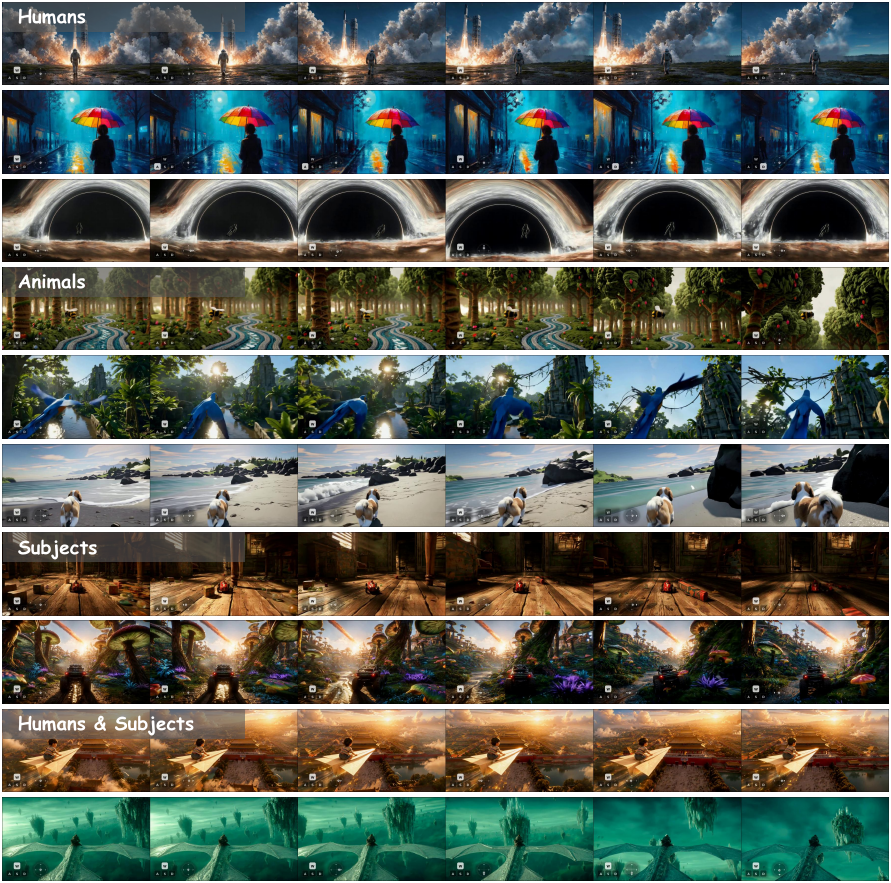}
    \caption{\textbf{Third-person generalization across visible subjects.}
    Representative rollouts span humans, animals, other foreground subjects, and scenes containing both humans and additional subjects.
    The shared camera-intent interface supports learned camera--character evolution across these subject configurations without a subject-specific controller.}
    \label{fig:qualitative-tpp}
\end{figure}

\noindent{\textbf{Third-Person Subject and Camera Control.}}
We next examine third-person interaction across different visible subject configurations.
Figure~\ref{fig:qualitative-tpp} groups rollouts containing humans, animals, other foreground subjects, and humans together with additional subjects.
Unlike an explicit actor trajectory, the condition specifies the desired evolution of observation; the learned world prior determines the associated subject motion and camera--character relationship.
This analysis therefore focuses on whether one shared interface produces coherent camera--character evolution across subject types without viewpoint- or embodiment-specific controllers.

\begin{figure}[t!]
    \centering
    \includegraphics[width=0.8\textwidth]{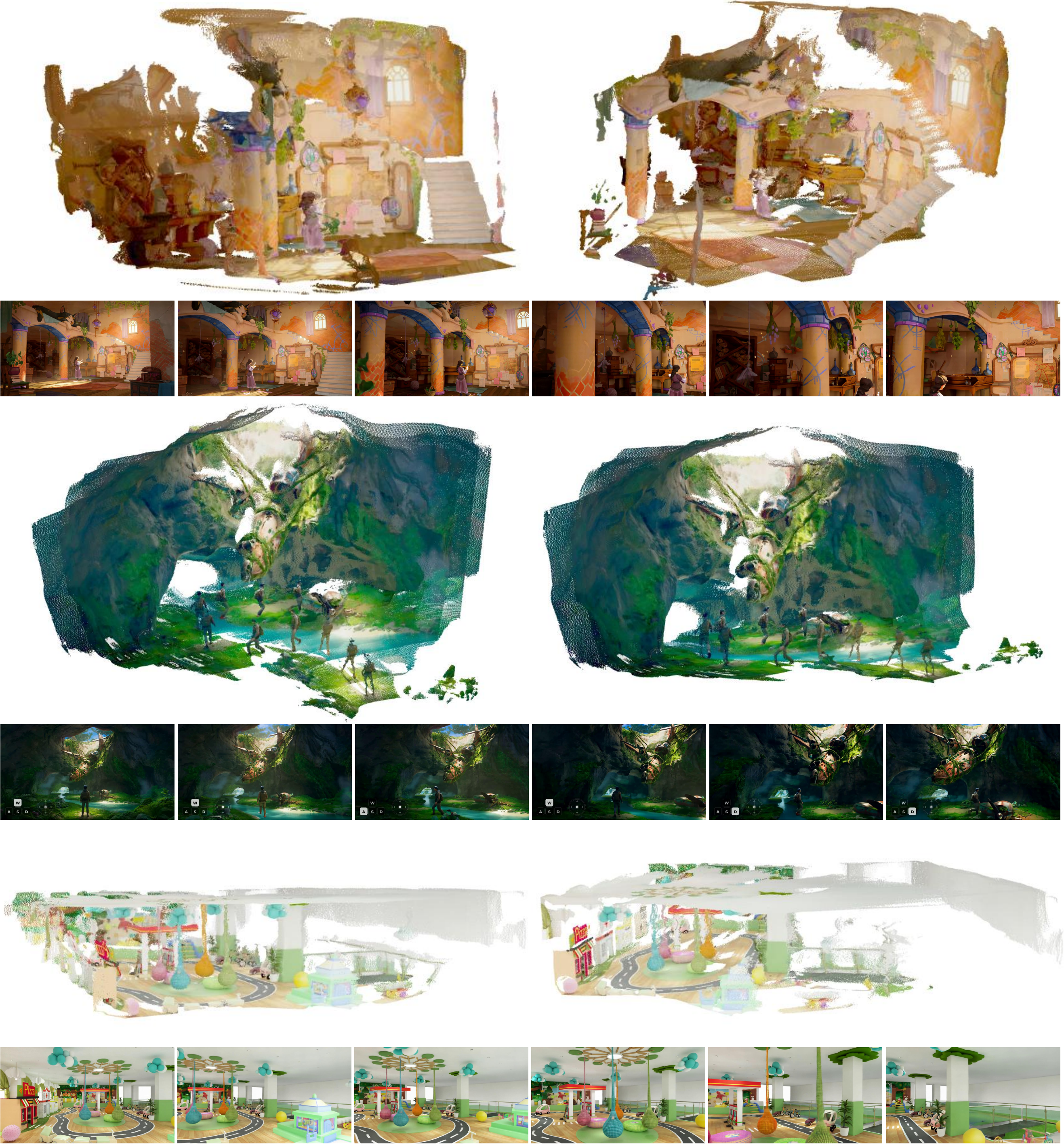}
    \caption{\textbf{Qualitative 3D reconstruction from generated trajectories.}
    Each example shows two views of the reconstructed scene together with six temporally ordered frames from the corresponding generated sequence.
    The recognizable scene-level layouts across viewpoint changes provide qualitative evidence of multi-view spatial coherence.}
    \label{fig:qualitative-3d-recon}
\end{figure}

\noindent{\textbf{Spatial Consistency through 3D Reconstruction.}}
Frame-wise visual quality alone does not establish that generated views describe a coherent 3D scene.
We therefore visualize 3D reconstructions obtained from generated sequences alongside the temporally ordered frames used for reconstruction in Figure~\ref{fig:qualitative-3d-recon}.
Across diverse indoor and outdoor settings, the reconstructed views retain recognizable scene-level layouts and major surfaces as the viewpoint changes.
Because the result also depends on the downstream reconstruction procedure, we treat it as qualitative supporting evidence of multi-view spatial coherence rather than a claim of metric geometric accuracy.

\noindent{\textbf{Generalization beyond the training domains.}}
Our training mixture contains no 2D or 2.5D game data and no robot data.
Figure~\ref{fig:generalization} evaluates the same camera-intent interface on these unseen domains: the top rows show 2D/2.5D game scenes, while the bottom rows show robot environments.
Across the displayed rollouts, \model maintains coherent viewpoint evolution and preserves major scene or subject structure under navigation controls.
This provides qualitative evidence of cross-domain generalization rather than evidence of domain-specific training coverage.

\begin{figure}[t!]
    \centering
    \includegraphics[width=0.9\textwidth]{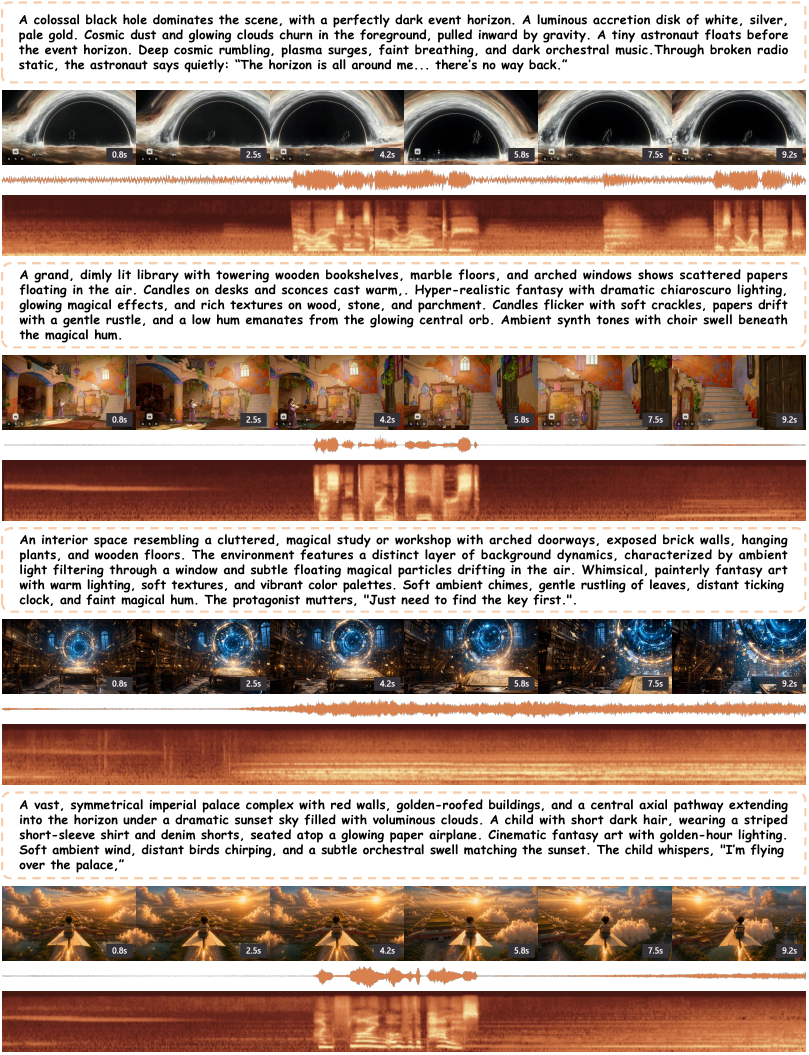}
    \caption{\textbf{Native synchronized audio-visual generation.}
    Each example pairs its world and audio description with time-stamped video frames, the generated waveform, and a log-frequency spectrogram.
    The cases contain environmental sound, background music, and character speech within the same generated audio-visual sequence, rather than adding audio as post-processing.}
    \label{fig:qualitative-av}
\end{figure}

\noindent{\textbf{Native Synchronized Audio-Visual Generation.}}
We then examine the model's native joint generation of video and audio across environmental sound, background music, and character speech.
Figure~\ref{fig:qualitative-av} pairs each structured world and audio description with time-stamped output frames, the generated waveform, and a log-frequency spectrogram.
The four examples include scene ambience and foreground effects, musical accompaniment, and spoken dialogue within the same generated sequences.
These examples assess synchronization between generated sound and visible content; they do not posit a separately supervised action-to-audio controller.

\begin{figure}[t!]
    \centering
    \includegraphics[width=1.0\textwidth]{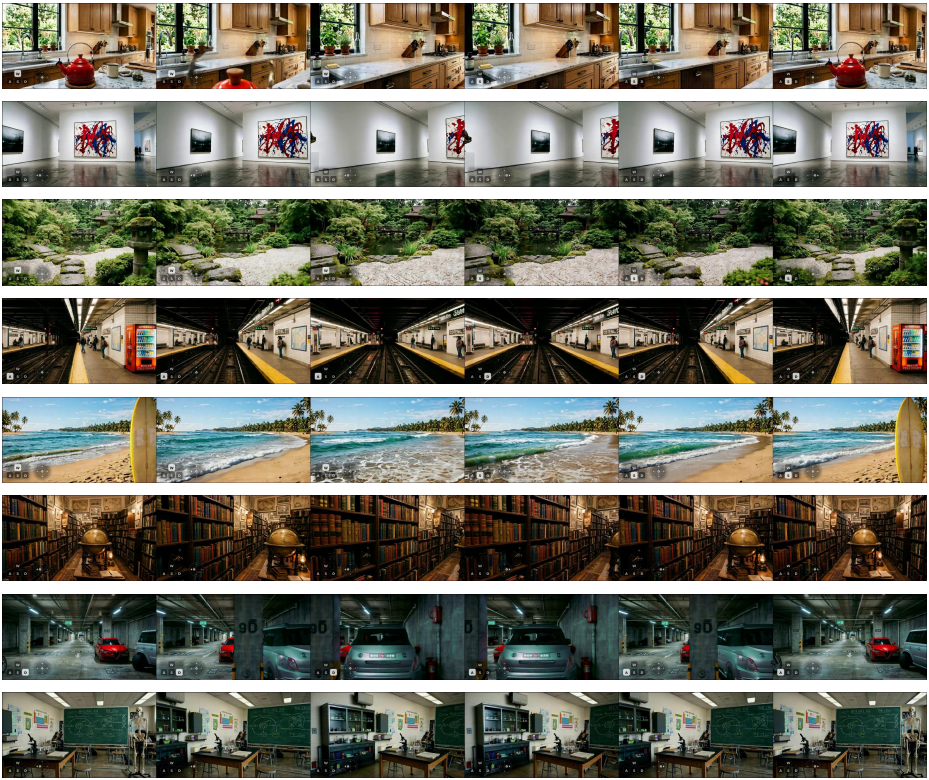}
    \caption{\textbf{Visual memory during long-horizon continuation.}
    Each row samples a temporally ordered trajectory that moves through and revisits a scene.
    Returning to previously observed views tests whether layout, appearance, and scene identity remain coherent over continued generation.}
    \label{fig:qualitative-multiturn}
\end{figure}

\noindent{\textbf{Multi-Turn Audio-Visual Continuation.}}
Finally, we examine whether continued generation preserves previously observed visual context.
Figure~\ref{fig:qualitative-multiturn} shows long-horizon trajectories that move through and revisit eight different scenes.
Each row samples the trajectory in temporal order, so the return to earlier viewpoints reveals whether layout, appearance, and scene identity remain coherent after intermediate motion.
This figure focuses on the visual-memory component of multi-turn continuation.

\noindent\textbf{Visual Memory.}
We assess whether the model preserves previously observed world states during
long-horizon continuation. Figure~\ref{fig:qualitative-multiturn} shows
trajectories that leave and later revisit earlier views. The returned views
retain major scene layout, appearance, and identity cues after intermediate
motion, providing qualitative evidence that the model can reuse prior visual
context across multiple turns.

\FloatBarrier

\subsection{Ablation Studies}
\label{sec:ablations}

We organize the ablations around trajectory scale, camera-intent semantics, inference-time speed control, and the audio-visual pretraining task mixture introduced in \cref{sec:method}.

\subsubsection{Trajectory Representation and Scale Calibration}
\label{sec:trajectory-design-eval}

The trajectory pathway must be both numerically stable during training and sensitive to displacement magnitude at inference.
We evaluate these properties separately because WBench adapts trajectory scale for robust open-domain scoring, which can hide errors in absolute motion magnitude.

\noindent{\textbf{Qualitative scale control.}}
We compare matched navigation rollouts to examine whether the fixed global
translation scale preserves the requested displacement amplitude as the camera
approaches a foreground character or salient scene structure.

\begin{figure}[ht]
    \centering
    \includegraphics[width=\textwidth]{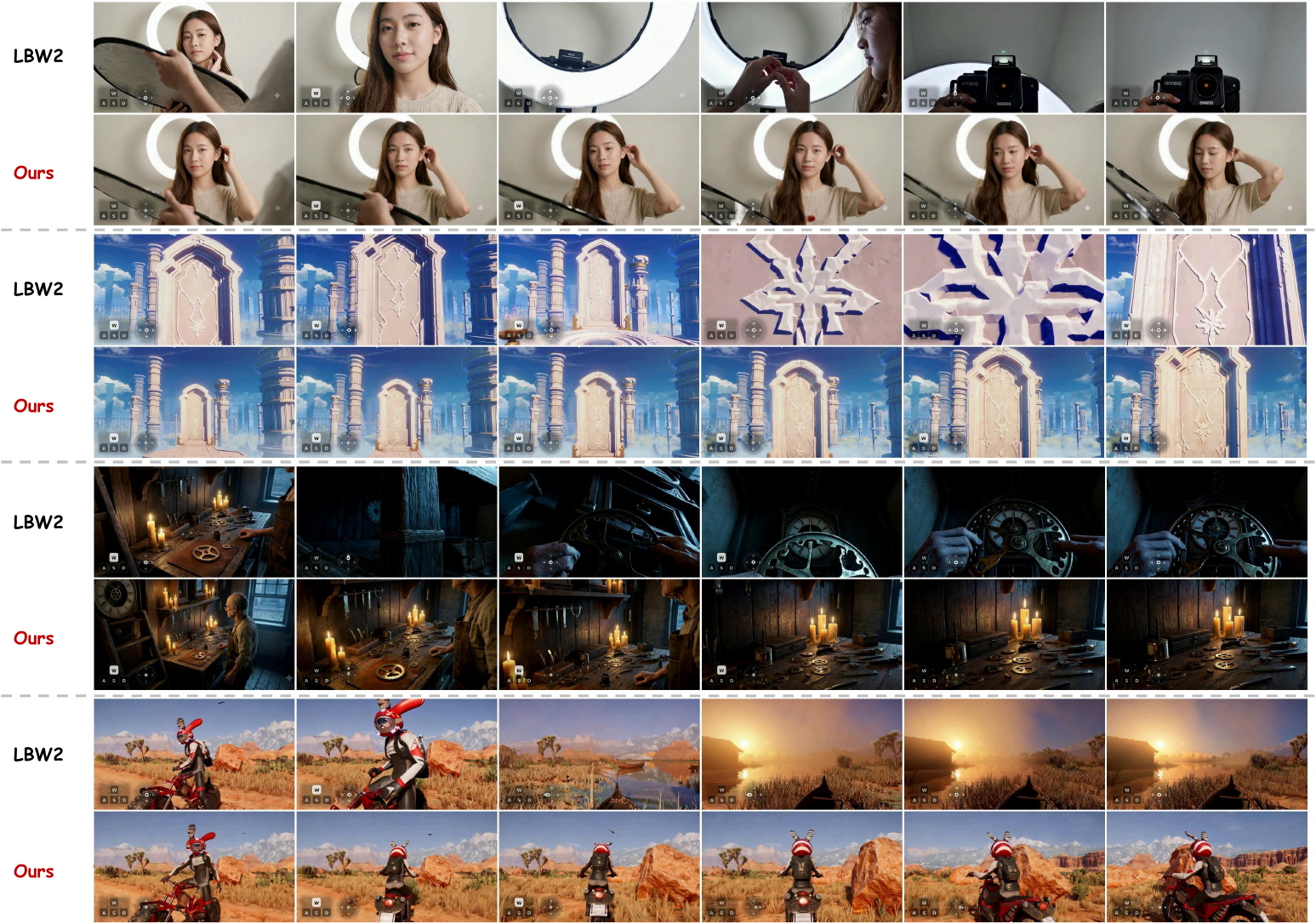}
    \caption{\textbf{Qualitative scale-control comparison.}
    LingBot-World-v2~\citep{gao2026lingbot} and \model receive matched reference observations and navigation conditions.
    When the camera approaches a foreground character or scene structure, LingBot-World-v2 exhibits larger changes in apparent subject scale and less predictable displacement response, whereas \model preserves a more gradual scale change across the trajectory.
    This qualitative comparison complements the metric-scale and speed-response ablations: a fixed global translation scale keeps requested displacement amplitudes comparable across clips and makes approach behavior more controllable.}
    \label{fig:metric-scale-compare}
\end{figure}

Figure~\ref{fig:metric-scale-compare} provides a qualitative complement to
this metric-scale ablation. Under matched navigation conditions, the
LingBot-World-v2 baseline can change apparent subject scale abruptly when the
camera approaches a character or a salient scene structure, making the final
distance and approach rate difficult to control. In contrast, the fixed global
translation scale used by \model preserves the relative amplitude of the
requested displacement across examples, yielding a more gradual and
predictable scale response. The figure is qualitative evidence for this
scale-control behavior; the inference-time speed analysis below is retained
for the corresponding quantitative evaluation.

\FloatBarrier

\noindent{\textbf{Inference-time speed control.}}
We construct trajectory groups that share the same reference observation, direction, and rotation but vary translation magnitude.
For each group, we measure the realized displacement in the generated video and report its correlation with the requested displacement, response slope, and absolute scale error.
The corresponding response curves test the central distinction between per-clip scaling, which removes cross-clip magnitude, and the fixed global scale in \cref{sec:pose-normalization}, which retains trajectory amplitude as a user control.
For multi-chunk rollouts, we also report the change in realized speed across chunk boundaries to test whether a normalization strategy introduces artificial transitions.

\begin{figure}[t!]
    \centering
    \includegraphics[width=\textwidth]{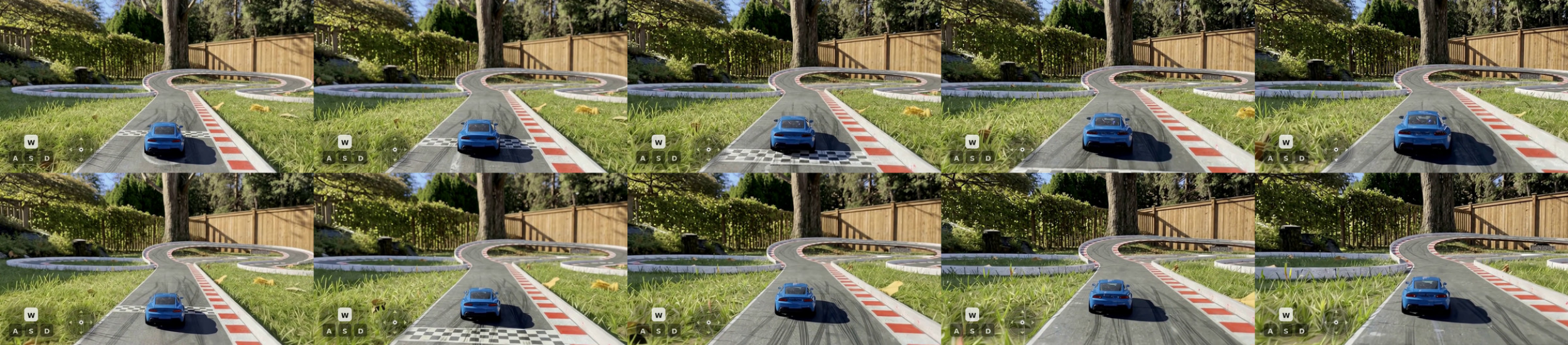}
    \caption{\textbf{Qualitative inference-time speed control.}
    Each row shows five temporally ordered frames from a navigation rollout
    under the same scene and control interface. The changing vehicle position
    illustrates the realized displacement over the rollout and complements the
    quantitative response curves described above.}
    \label{fig:inference-speed-control}
\end{figure}

\subsubsection{Camera-Intent Modeling across Viewpoints}

We qualitatively examine whether one camera-intent interface supports direct
observer motion in first person and coordinated camera--character evolution in
third person without viewpoint-specific controllers.

\begin{figure}[t!]
    \centering
    \includegraphics[width=\textwidth]{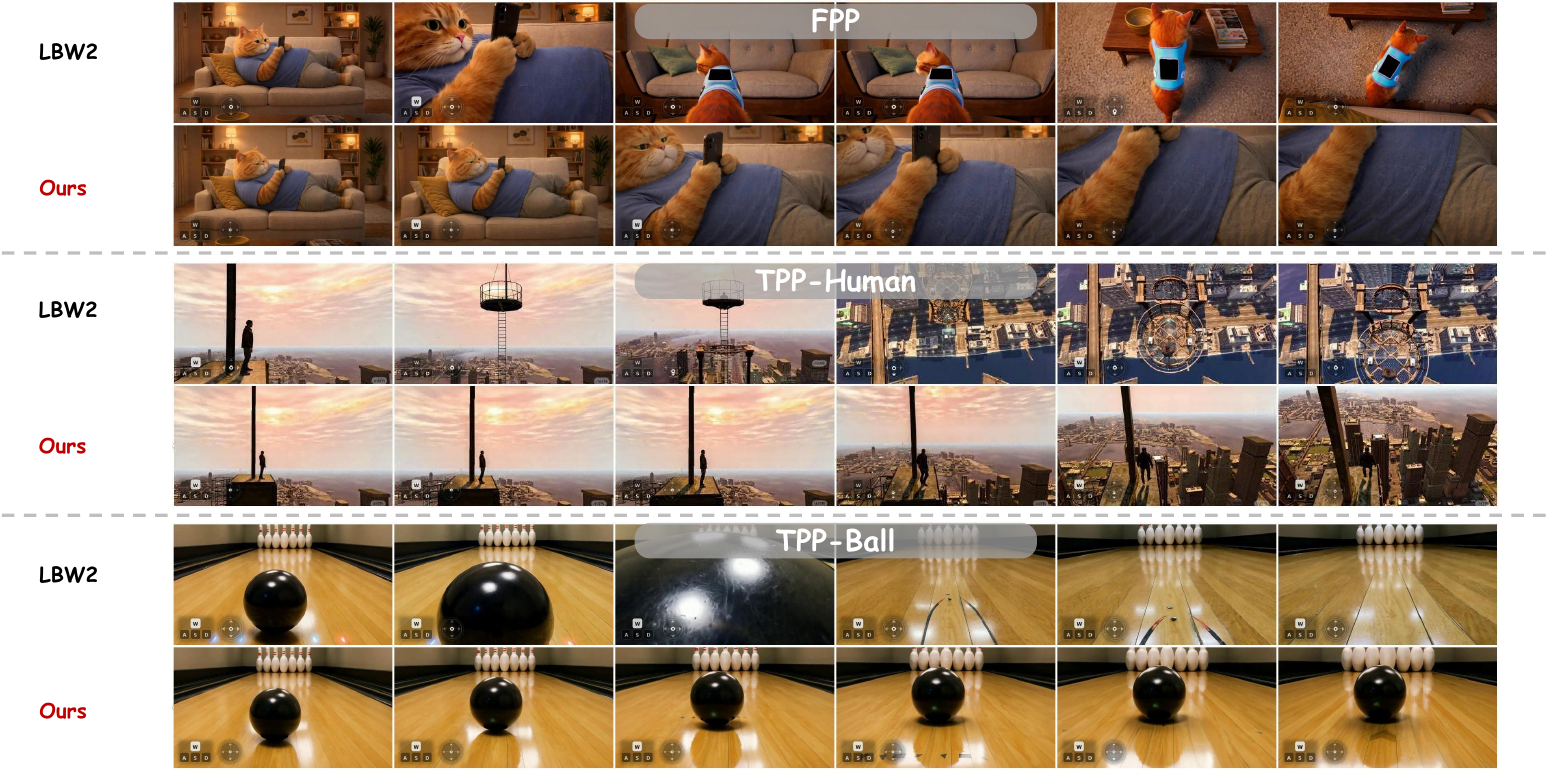}
    \caption{\textbf{Qualitative viewpoint semantics and third-person tracking.}
    LingBot-World-v2 (LBW2)~\citep{gao2026lingbot} and \model receive matched reference observations and navigation conditions in first-person (FPP), third-person human (TPP-Human), and third-person object (TPP-Ball) cases.
    In the FPP example, LBW2 changes to an exocentric observation as the rollout proceeds, whereas \model retains the requested first-person semantics.
    In both TPP examples, LBW2 loses the controlled subject from view, while \model keeps the human or ball visible and maintains a more stable camera--subject relationship across the sequence.}
    \label{fig:viewpoint-tracking-compare}
\end{figure}

Figure~\ref{fig:viewpoint-tracking-compare} provides a qualitative comparison
with LingBot-World-v2 under matched first- and third-person controls. In the
first-person case, \model preserves the requested egocentric observation and
avoids the transition to an exocentric view observed in the baseline rollout.
In the two third-person cases, \model also keeps the human or ball visible
throughout the sequence and maintains a more stable camera--subject
relationship, whereas the baseline loses the subject as the camera evolves.
These examples provide qualitative evidence for the role of explicit viewpoint
semantics and a shared camera-intent condition.



\FloatBarrier
\section{Discussion}
\label{sec:discussion}

\model connects two directions that have largely developed separately: generative audio-visual media and interactive world models.
Its design suggests five principles that extend beyond the particular backbone.

\noindent{\textbf{Center interaction on observation intent.}}
Rather than defining a separate control semantics for each viewpoint, we describe how the user intends to observe and traverse the world.
The same camera intent can be realized as observer motion in first-person scenes or as coordinated camera and character evolution in third-person scenes.
This lets the model learn plausible camera--character coupling from data, while cinematic video and gameplay contribute complementary evidence about observation, composition, and world response.

\noindent{\textbf{Represent interaction at the world level.}}
Input devices and benchmarks express navigation through different action vocabularies, but their effect can often be represented as a common camera-space change.
Converting discrete and continuous inputs to a shared relative trajectory avoids binding the generator to a particular input layout or action encoder.
This observation applies to navigation and viewpoint-related actions; it does not imply that arbitrary actor behavior can be reduced to camera motion.

\noindent{\textbf{Make scale and viewpoint semantics explicit.}}
Per-example normalization can make optimization easier while removing the relative displacement that an interaction condition is intended to represent.
A robust dataset-level scale preserves motion relationships across sources and keeps adjacent rollout chunks on the same velocity scale, avoiding normalization-induced speed jumps.
At the same time, geometry alone does not specify whether a rotation is egocentric, character-following, or orbital.
Conditioning on viewpoint separates camera intent from its scene-dependent realization.

\noindent{\textbf{Keep audio inside the generated world.}}
In passive media, audio enriches presentation; in an enterable world, it also provides ambient, off-screen, and social context while the visual rollout is being navigated.
Our trajectory condition enters only the video stream, so we do not impose an explicit action-to-sound controller.
Instead, environmental sound and speech remain part of the native joint audio-visual process.
Joint-FT optimizes this process on trajectory-conditioned examples, but the trajectory still reaches audio only through the model's existing cross-modal coupling.
We therefore evaluate acoustic coherence under interaction without claiming a separately supervised action-to-sound capability.

\noindent{\textbf{Match each training stage to its cleanest signal.}}
Audio-visual generation and interaction rely on different data distributions and objectives.
AV-CPT uses audio-rich and speech-rich clips to establish the world prior, while Action-SFT freezes that prior and uses control-clean motion to isolate navigation acquisition.
A smaller balanced set then supports low-learning-rate Joint-FT of the full model.
The curriculum thus separates capability acquisition before allowing the two pathways to co-adapt.

Together, these principles explain why \model is more than a controlled video generator.
The data engine supplies complementary world experience, the shared interaction condition makes it enterable, the joint backbone makes it perceptually responsive, and continuation turns individual generations into an evolving audio-visual experience.

\section{Limitations}
\label{sec:limitations}

\model currently focuses on navigation and viewpoint-related actions that can be represented by a relative 6-DoF trajectory.
Discrete keyboard states are supported through this mapping, but the model does not explicitly represent arbitrary actor intent such as jumping, attacking, manipulation, or robot commands.
It also does not implement the rules, collision state, or deterministic transitions of a conventional game engine.

The current model lacks an explicit persistent 3D memory.
Geometry, subject identity, world state, and audio can therefore drift over repeated continuation turns.
Its interaction range is also restricted to trajectories retained by the global filter; generalization to substantially larger translations, higher speeds, or unusual rotations is not established.
Pose estimates from Internet and gameplay video remain imperfect even after filtering and may correlate with scene content or motion blur.

\section{Conclusion}
\label{sec:conclusion}

We presented \model, an omnimodal world model for enterable generative media that bridges generative media and interactive world modeling across general and game scenes.
\model organizes interaction around camera intent, allowing first-person observer motion and third-person camera--character evolution to be learned as different realizations of how a world is observed.
It combines this perspective with a complementary gameplay--simulation--Internet data engine, a scale-consistent trajectory condition, and a progressive audio-visual-to-action curriculum.
Continued audio-visual pretraining first learns from AV-rich data; action fine-tuning freezes this world prior and learns visual navigation from control-clean motion; low-learning-rate joint fine-tuning then consolidates both on a smaller, jointly high-quality set.
Sound remains part of the jointly generated world and provides perceptual context during interaction.
The resulting model supports first- and third-person interaction, native joint audio-visual output, and multi-turn continuation in one system.
More broadly, our results suggest a path from generative content that is only watched toward omnimodal worlds that can be continuously entered and influenced.

\clearpage
\bibliographystyle{plainnat}
\bibliography{cite}


\end{document}